\pdfoutput=1

\documentclass[11pt]{article}

\usepackage[final]{acl}

\usepackage{times}
\usepackage{latexsym}
\usepackage{algorithmic}
\usepackage{textcomp}
\usepackage{xcolor}
\usepackage{multirow}
\usepackage{multicol}
\usepackage{tabularx}
\usepackage{comment}
\usepackage{color}
\usepackage{hhline}
\usepackage[normalem]{ulem}
\usepackage{bm}
\usepackage{booktabs}
\usepackage{amsmath}
\usepackage{enumitem}
\usepackage[export]{adjustbox}
\usepackage{rotating}
\usepackage{booktabs}
\usepackage{caption}

 \usepackage{tabularray}

\usepackage[most]{tcolorbox} 

\definecolor{denim}{rgb}{0.08, 0.38, 0.74}
\definecolor{feldgrau}{rgb}{0.3, 0.36, 0.33}
\definecolor{greenmun}{rgb}{0.0, 0.66, 0.47}
\definecolor{jade}{rgb}{0.0, 0.66, 0.42}
\definecolor{lightsalmonpink}{rgb}{1.0, 0.6, 0.6}
\definecolor{lightcoral}{rgb}{0.94, 0.5, 0.5}
\definecolor{cardinal}{rgb}{0.77, 0.12, 0.23}
\definecolor{carnelian}{rgb}{0.7, 0.11, 0.11}
\definecolor{greenncs}{rgb}{0.0, 0.62, 0.42}

\usepackage[T1]{fontenc}

\usepackage[utf8]{inputenc}

\usepackage{microtype}

\usepackage{inconsolata}

\usepackage{graphicx}

\title{Evaluating LLMs on Large-Scale Graph Property Estimation \\ via Random Walks}

\author{Sunil Kumar Maurya \\
  University of Tokyo \\
  Tokyo, Japan \\
  \texttt{sunil.maurya@weblab.t.u-tokyo.ac.jp} \\\And
  Xin Liu\thanks{~~Corresponding Author.} \\
  AIST \\
  Tokyo, Japan \\
  \texttt{xin.liu@aist.go.jp} \\}

\begin{document}
\maketitle
\begin{abstract}
With the rapidly improving reasoning abilities of Large Language Models (LLMs), there is also a rising demand to use them in a wide variety of domains. This brings about the need to carefully evaluate the limits of the capabilities of these models with various tests and benchmarks. Graph structures are ubiquitous in real-world data, and are often used to represent and analyze relationship patterns within data. Many benchmarks have already been proposed in the graph literature to test the reasoning ability of LLMs to follow and execute graph algorithms. However, due to the limited context length of LLMs, these benchmarks consist of very small graphs. In real-world data, the size of graphs can be significantly larger, and in many cases, not fully accessible. \textcolor{black}{In this paper, we examine a class of problems that arises with very large graphs having limited accessibility. We propose a large graph benchmark dataset, \texttt{EstGraph}, and introduce four distinct tasks designed to estimate large graph properties. We evaluate the reasoning abilities of LLMs on these tasks using a wide variety of graph datasets. In addition, we provide task-specific prompt constructions based on random walk sampling of large graphs (up to millions of nodes)  that effectively convey sufficient information to LLMs within the limits of context length. Source code and datasets are available at \url{https://zenodo.org/records/19632942}}

\end{abstract}

\section{Introduction}

Large Language Models (LLMs) with reasoning abilities have opened up opportunities to utilize human-like thinking process for a wide variety of tasks and scientific domains. However, the reasoning abilities of these models are still limited. In addition, they suffer from hallucinations and can often make mistakes during multi-step operations. Therefore, researchers are making considerable efforts to benchmark the capabilities of LLMs, and new releases of LLMs are often accompanied by evaluations on a range of benchmarks to showcase their knowledge and reasoning capabilities \cite{cao_toward_2025}.

Similarly, there have been extensive efforts in the field of graph theory to propose new benchmarks that measure the reasoning abilities of LLMs on graph-related algorithmic tasks. Some of these benchmarks are NLGraph \cite{wang_can_2023}, GraphQA \cite{fatemi_talk_2024}, GraphArena \cite{tang_grapharena_2025}, GraphPattern \cite{dai_how_2025}, and so on. These works explore the ability of LLMs to parse graph structures described in natural language and solve graph problems. These benchmarks include questions that explore topics of local structural understanding in graphs, such as node connectivity and neighborhood identification, and global reasoning tasks, including finding shortest paths, Hamiltonian paths, or graph diameter.

\begin{figure}
    \centering
    \includegraphics[width=0.90\linewidth]{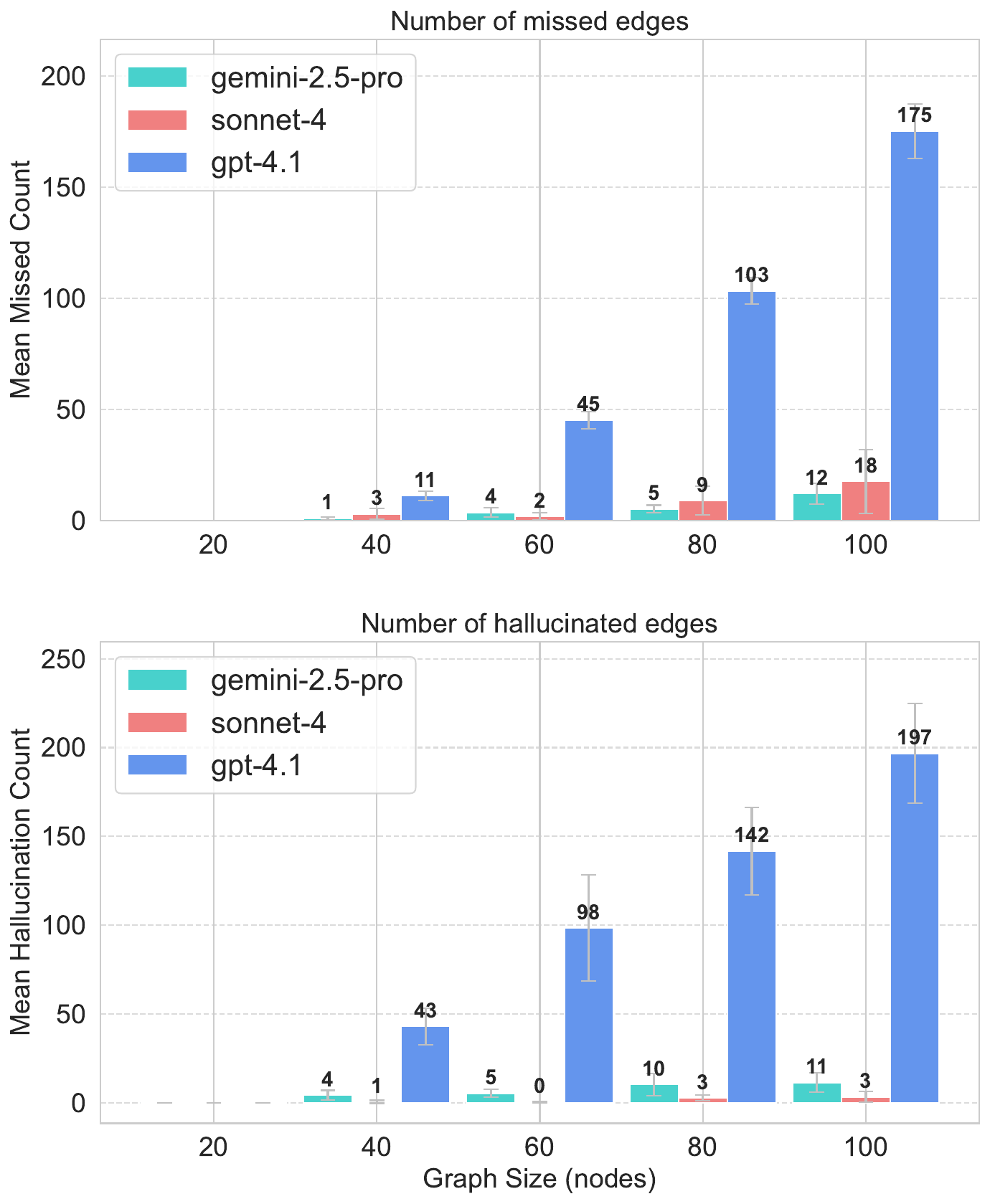}
    \caption{Plot shows number of missed edges (mean of 5 graphs for each size) and number of hallucinated edges with increase in the graph size when converting from edgelist to adjacency list for three LLMs\protect\footnotemark.}
    \label{fig:g_conversion_plot}
\end{figure}

Evaluations on these benchmarks show that LLMs perform well on small graphs and demonstrate graph reasoning abilities. However, as the size of the graphs grows larger, they tend to perform worse and have difficulty maintaining the context of the global structure of the graph. Figure \ref{fig:g_conversion_plot} shows the performance of LLMs on a simple task of conversion from an edgelist (tuples of edges) of a graph to an adjacency list (dictionary of node as keys and list of neighbors as values). The task only requires maintaining the context of local neighborhood. However, as the size of the graphs grows, all LLMs start to miss existing edges or hallucinate new edges that do not exist in the graph. Reasoning models have fewer errors than non-reasoning models, nevertheless, the number of errors increases with the graph size.  \cite{fatemi_talk_2024} tested various graph encoding methods for LLMs and found that even for the same graph and same graph problem, LLM performance can vary significantly based on how the graph is encoded in the input prompt \cite{wang_can_2023,finkelshtein_actions_2026}.
\footnotetext{ We used OpenAI's gpt-4.1 instead of Reasoning o3 model, as it frequently refused or timed out in this experiment.}
While these recent works have significantly improved our understanding of the capabilities of LLMs on graph problems, the graph sizes used in the benchmarks remain small, with only 20-50 nodes in each graph (Table \ref{tab:benchmark-comparisons}). \textcolor{black}{This limitation stems from two factors: context limits can prevent full edge information from fitting in the prompt, and even if it fits, LLMs often struggle to maintain a consistent global view of large graph structure, increasing errors.}

Due to these limitations, benchmarks related to reasoning on graph structures consist of small-sized graphs, and many problem domains that require reasoning over large graphs remain underexplored. One such research area is estimation of properties of large graphs that are not fully accessible. \textcolor{black}{Many real-world graphs (e.g., social, peer-to-peer, and web graphs) can have thousands to millions of nodes and are often only partially accessible via APIs. As graphs scale, attention shifts from fine-grained algorithmic details to global structure, such as community organization, connectivity patterns, and degree distributions, which are useful for analyzing information diffusion and robustness to failures or targeted attacks. Extensive research has studied estimating graph properties such as the numbers of nodes/edges, cliques, and clustering coefficients from sampled subgraphs. \cite{gjoka_practical_2011, katzir_estimating_2011, massoulie_peer_2006, hardiman_estimating_2013}.}

\textcolor{black}{We structure this study around the following research questions:}
\begin{itemize}[itemsep=-2mm]
\color{black}
    \item How accurately can LLMs estimate global properties of large graphs using only random-walk statistics?
    \item How do LLMs compare with classical estimators across diverse graph structures and tasks?
\end{itemize}

Following are the contributions of our work:

\begin{enumerate}[label=(\roman*)]
    \item  \textcolor{black}{We propose a large graph benchmark \texttt{EstGraph} and introduce four tasks to estimate large-graph properties that are relevant to real-world applications.}

    \item \textcolor{black}{We propose task-specific prompt constructions that encode the statistics about graph structure and avoid limitations related to context length. For the task of graph size estimation on real-world datasets, our approach reduces prompt length by up to 559x.}

    \item \textcolor{black}{We evaluate reasoning LLMs on graph-related estimation tasks in large graphs with up to millions of nodes, and in a real-world setting with limited access to the graphs.}

    \item \textcolor{black}{Based on our experiments, we also provide practical recommendations for large graph analysis with LLMs (Section \ref{sec:discussion} \& \ref{sec:recommendation}).}
\end{enumerate}

\begin{table*}[t]
\centering
\resizebox{\linewidth}{!}{%
\begin{tabular}{l c c c c}
\toprule
\textbf{Benchmark} &
\textbf{Scale (nodes)} &
\textbf{Access assumption} &
\textbf{Task goal} &
\textbf{Graph Encoding} \\
\midrule
NLGraph~\cite{wang_can_2023} &
Max: 20 &
Full graph &
Algorithmic execution &
Edgelist text \\
GraphQA~\cite{fatemi_talk_2024} &
Avg: 12.3 &
Full graph &
Algorithmic execution &
Edgelist / adjacency text \\
GraphPattern~\cite{dai_how_2025} &
Avg: 29.5 &
Full graph &
Algorithmic execution &
Edgelist / adjacency text \\
GraphArena~\cite{tang_grapharena_2025} &
Max: 50 &
Full graph &
Algorithmic execution &
Edgelist text \\
\midrule
\textbf{EstGraph (ours)} &
\begin{tabular}[c]{@{}c@{}}
Max: 100k (synthetic) \\
Max: 2.39M (real-world)
\end{tabular} &
Partial graph &
Estimation under partial access &
Random-walk statistics \\
\bottomrule
\end{tabular}}
\caption{\textcolor{black}{Comparison of existing LLM-reasoning on graph benchmarks with EstGraph}}
\label{tab:benchmark-comparisons}

\end{table*}

\section{\textcolor{black}{Related Works}}

 \textcolor{black}{With rapid improvements in LLMs, various benchmarks have been proposed to evaluate their reasoning abilities over graph-structured data. Benchmarks like GraphQA \cite{fatemi_talk_2024}, NLGraph \cite{wang_can_2023} and GraphArena \cite{tang_grapharena_2025} evaluate structure-based graph reasoning, posing questions that require inferring solutions from a graph’s topology, and share many tasks, like adjacency/neighborhood queries, connectivity or cycle detection, finding shortest paths, and so on. In contrast, GraphPattern \cite{dai_how_2025} evaluates the LLMs ability to recognize primitive patterns or small-subgraph motif structures.} 

 \textcolor{black}{In these benchmarks, tasks assume a full-visibility setting, where the entire graph information is provided as a textual representation in the prompt. An LLM is then asked to produce an exact answer to a well-defined query. This formulation is valuable for testing algorithmic reasoning on small graphs, but it becomes infeasible as the graph size grows. The sizes of graphs in these benchmarks are limited to at most 10-50 nodes (Table \ref{tab:benchmark-comparisons}). As a consequence, existing evaluations provide limited insight into how LLMs behave when the graph is too large to be shown and only partial information can be obtained.}

\textcolor{black}{Random walks are used for exploring graph structure using only local neighborhood access and provide a scalable probe. They are often used as a tool for search on graphs \cite{bisnik_modeling_2005, bockling_walkretrieve_2025}, ranking \cite{brin_anatomy_1998}, node embedding \cite{perozzi_deepwalk_2014, grover_node2vec_2016}, or walk-based graph representational learning \cite{kim_revisiting_2025, ivanov_anonymous_2018}.}

\textcolor{black}{\textbf{Positioning of EstGraph}: Existing LLM graph reasoning benchmarks primarily test exact question answering under full visibility on small graphs, whereas many practical large-graph settings require estimations from partial or limited access. Our work introduces estimation tasks over graphs with limited access and enables evaluation of LLM reasoning on graphs that are orders of magnitude larger via random walk sampling of graphs. Thus, EstGraph complements the prior graph-reasoning benchmarks and expands the evaluation to the large-scale regime.}

\begin{figure*}[h]
    \centering
    \includegraphics[width=0.85\textwidth]{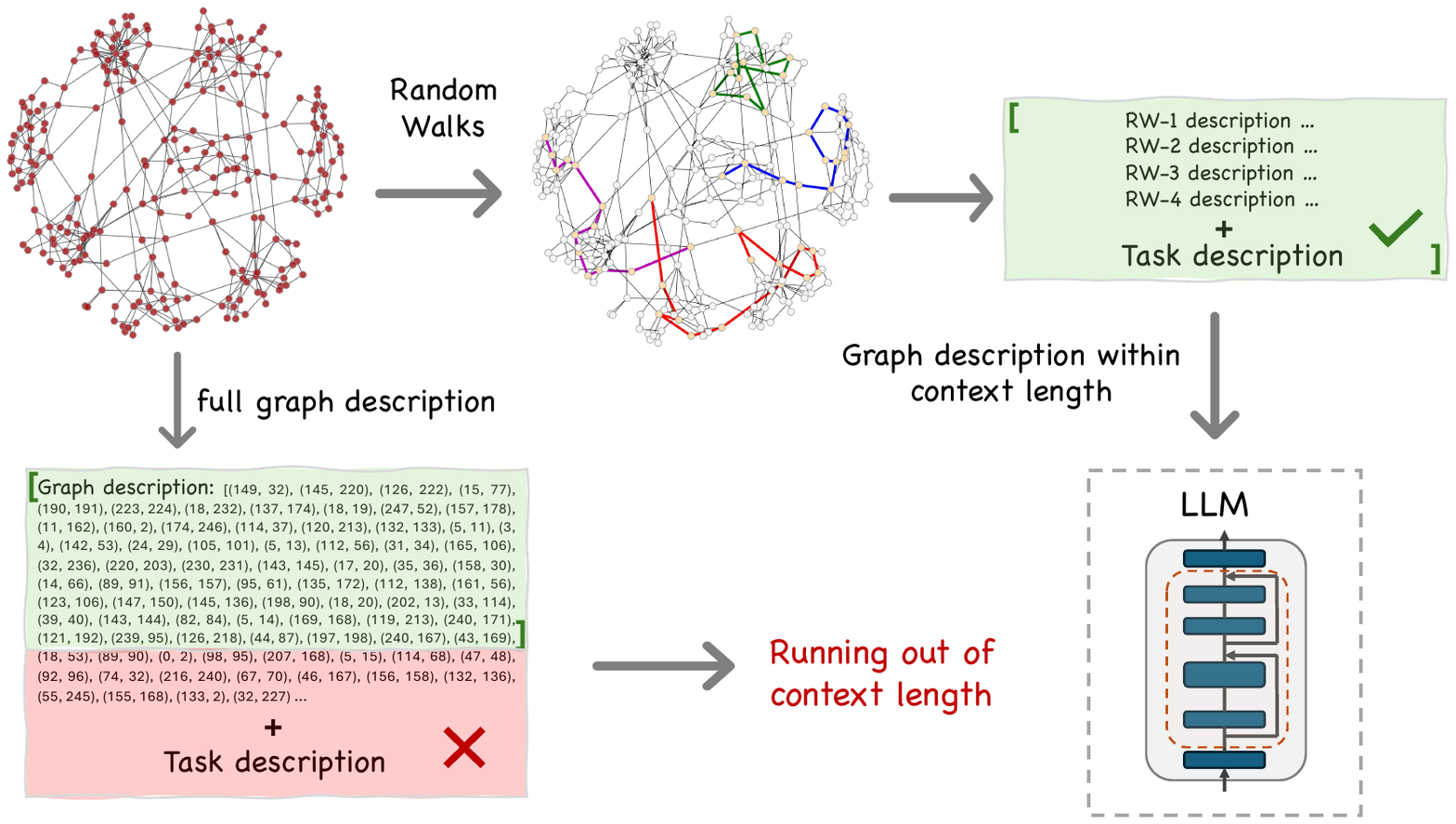}
    \caption{Figure illustrates the issue of exceeding context length as the graph size increases. Random walks on graphs provide efficient way of extracting and encoding graph-related information. }

    \label{fig:main_figure_rw}
\end{figure*}

\section{Preliminaries}

\subsection{Definitions}
Consider an undirected graph $G = (V,E)$ with $n = |V|$ nodes and $m = |E|$ edges. The neighborhood $\Gamma(u)$ of a node $u$ is the set of nodes $v$ such that ${u,v}\in E$. The degree of a node $u$ is 

\[d(u) := |\Gamma(u)| = |\{v  |  \{u,v\} \in E\}|\]

and the average degree of a node $u$ is

\[d_{avg} := \frac{1}{n} \sum_{u}d(u) = \frac{2m}{n} \]

Encoding the entire graph structure in a prompt is a difficult task, especially for large graphs. However, we can sample from a large graph in such a way that the resulting subgraph still provides sufficient information for the task while satisfying the context length limitations of LLMs. In the next section, we discuss the details of using random walks as an approach to sample nodes from large graphs.

\subsection{Random Walks}
Random walks on graphs are often used to sample nodes in graph mining applications, following connectivity patterns of the nodes. While at the current node $u$, the characteristics of the random walk are determined by how the next node $v$ is chosen. For a \textit{simple random walk}, a node $v$ is picked uniformly at random from $\Gamma(u)$, and the walk moves to node $v$. The transition matrix $P^{srw}$ associated with this random walk on $G$ is given by
\[
P^{\text{srw}}(u,v) = 
\begin{cases}
    \frac{1}{d(u)} & \text{if } \{u,v\} \in E, \\
    0 & \text{otherwise},
\end{cases}
\]

 The distribution of nodes and their degrees in a random walk encodes characteristics of the graph. However, random walks need to cover a sufficient portion of the graph to provide an accurate representation of the graph. Therefore, as the graph size increases, the amount of random walk data also increases linearly. As a result, directly providing random walk data in the prompt faces limitations due to the context length of LLMs. Instead, depending on the task, we can summarize the statistics of the walks that can easily scale with the graph size.

\section{Estimation of Graph Properties}
\label{sec:est_tasks}

\subsection{Estimation of Number of Nodes and Edges}
\label{sec:est_node_edge}
Estimating the size of a graph and other properties has been extensively studied in the graph literature especially for social networks like Facebook or Twitter. One common way is to use a random walk as a surrogate for uniform sampling of nodes. After sampling, we can use either \textit{sample and collide} method which relies on the same phenomenon as the "birthday paradox" \cite{katzir_estimating_2011, chiericetti_sampling_2016} or use the \textit{capture-recapture} method that uses the Chapman Estimator (a corrected Lincoln-Peterson estimator) \cite{accettura_capture-recapture_2015}. Both methods are similar, and in this work, we utilize Chapman Estimator to estimate the graph size $\Hat{N}$,

\begin{equation}
    \Hat{N} = \frac{(|\mathcal{S}_1| + 1)(|\mathcal{S}_2| + 1)}{|\mathcal{C}|} - 1
\end{equation}

where $\mathcal{S}_1$ and $\mathcal{S}_2$ are two independent sets of nodes sampled with unbiased random walks, and $|\mathcal{C}|$ is the number of common nodes in both sets.

Using the estimator, we can estimate the number of nodes by uniformly sampling two sets of nodes from the graph. To estimate the number of edges, we can find the average degree of the sampled nodes and calculate number of edges as $\Hat{M}=\frac{d_{avg}\cdot\Hat{N}}{2}$. The following are some methods in the research literature that are often used to uniformly sample large graphs and estimate their properties.

\textit{Uniform Sampling from Nodelist} As our first baseline, we take a simple approach of uniformly sampling nodes from the nodelist of the graph. 

\textit{Metropolis-Hastings sampling} This method uses a modified random walk to draw nodes from the graph such that the transition probabilities of the walk converges into a uniform distribution \cite{metropolis_equation_1953,gjoka_practical_2011}.

\textit{Max-degree Random Walk Sampling} In this algorithm, to uniformly sample nodes via random walk, the original graph is modified into a regular graph (where all nodes have same degree) by adding self-loops to the nodes so that the degree of each node equals the maximum degree ($d_{max}$) of the original graph \cite{bar-yossef_approximating_2000,li_random_2015}.

\textit{Return-time Based Estimation} This method estimates global properties of a graph using the times of the first returns of random walks \cite{cooper_fast_2016} . 

In practice, many real-world graphs such as online social networks are not fully accessible. Often uniform sampling based methods are not efficient as they either require rejecting samples to avoid bias in sampling or require global knowledge such as the node list or maximum degree, which is not feasible in such settings.

\subsection{Estimation of Number of Communities}
Real-world graph data often consists of clusters or communities that can arise due to social dynamics, geography, or some node interaction patterns depending upon the source of the data. In such graphs, a node has higher probability of edge existing between another node within same cluster than with a node from outside the cluster. This task involves detecting the number of clusters or communities in a given graph.

To benchmark community detection algorithms, we use Lancichinetti-Fortunato-Radicchi (LFR) algorithm \cite{lancichinetti_benchmark_2008} to generate synthetic graphs with a \textit{priori} known communities and use three community detection algorithms as baselines: \textit{Greedy Modularity} \cite{clauset_finding_2004}, \textit{Louvain Community Detection}, \cite{blondel_fast_2008} and \textit{Label Propagation} \cite{cordasco_community_2010}.

In our experiment to compare the performance with LLMs, instead of using the full graph, we use subgraphs induced by random walks on graphs to detect number of communities.

\begin{figure}
    \centering
    \includegraphics[width=0.83\linewidth]{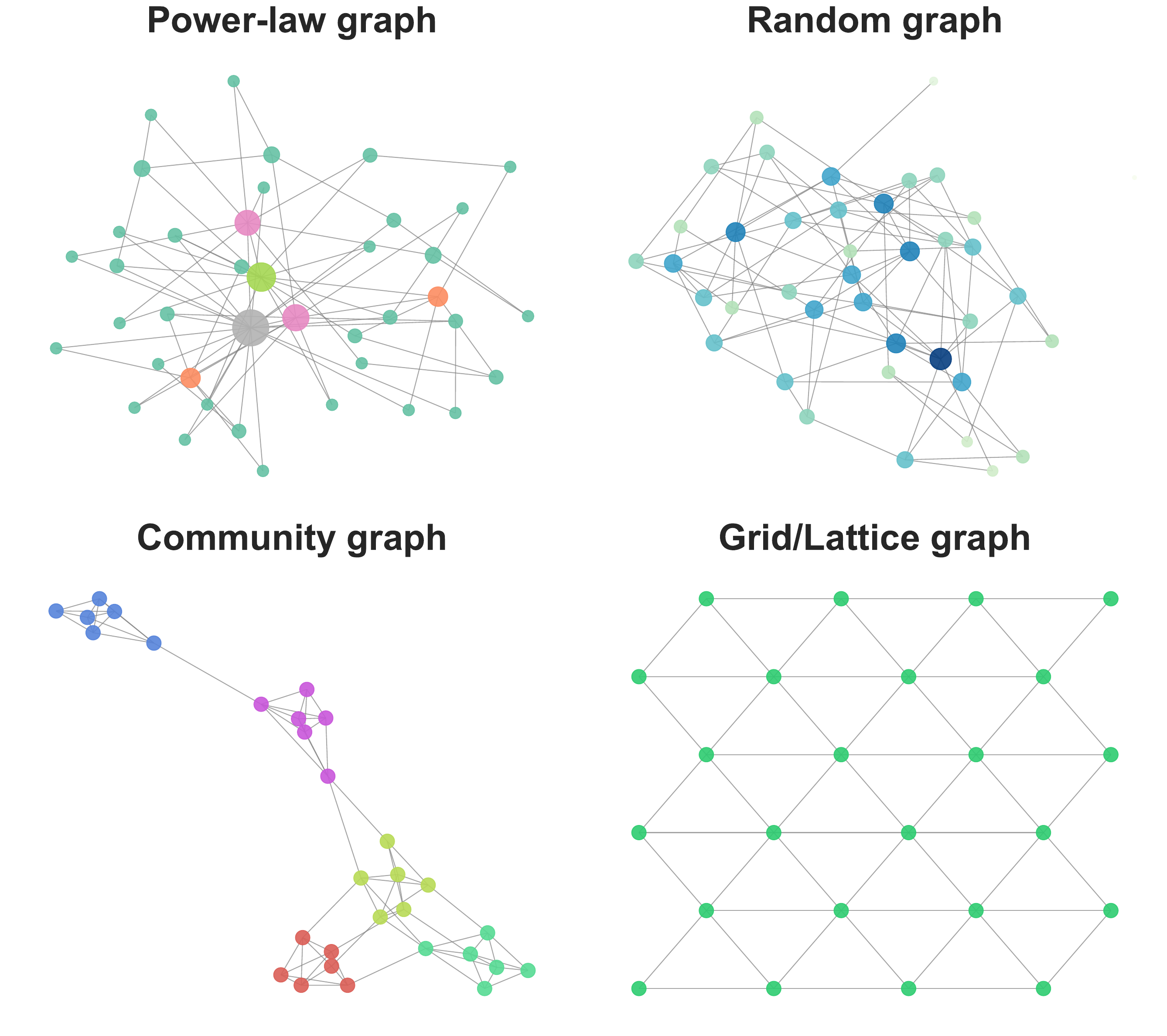}
    \caption{Graphs with varied structures}

    \label{fig:graph_types}
\end{figure}

\subsection{Estimation of Graph Structure}
Real-world graphs can often exhibit a wide range of structural properties depending on the domain. These graphs are often studied to extract connectivity patterns and understand network dynamics \cite{yan_efficient_2006, leskovec_planetary-scale_2008}. Random walks can be used to extract such information from the graphs \cite{rosvall_maps_2008}. In this task, we evaluate the LLMs' capability to identify the structure of the graph by providing the degree distribution of the nodes visited in multiple random walks.

\subsection{Estimation of Influential Nodes in the Graph}

One of the important applications of graphs in real-world data is the identification of influential nodes in a network.
These nodes often affect network dynamics by acting as critical nodes for information spread (social networks, computer networks, biological networks, etc.) or as intersections or bottlenecks (transportation networks, supply-chain networks, etc.) \cite{newman_networks:_2010, brin_anatomy_1998}. In this task, we test the ability of LLMs to identify node ranking measures such as \textit{Betweenness Centrality}, \textit{Closeness Centrality}, and \textit{PageRank} using simple random walks. These measures are typically calculated on full graphs and are challenging to compute with partial access to the graphs.

\section{Experiments}
In this section, we provide details of the experimental settings with baseline graph algorithms and LLMs.
\subsection{Datasets}
For the task of estimating the number of nodes and edges, we generated synthetic graphs of three types: Barabasi-Albert (BA) or power-law graphs, Erdos-Renyi (ER) random graphs, and Gaussian Random Partition (GRP) random graphs with clusters. In order to evaluate performance variations with graph scale, we divided the graphs into three categories: Small (100-1,000 nodes), Medium (1,000-10,000 nodes) and Large (10,000-100,000 nodes). In addition, we evaluated the LLMs on five real-world datasets from the SNAP graph repository, with number of nodes ranging up to millions.

For the second task of estimating the number of clusters in a graph, we used 20 LFR synthetic graphs that are often used to benchmark community detection algorithms. The number of clusters in the graphs ranges from 5 to 12.

The third task measures LLMs' capabilities to identify the basic structure of the graph. For this task, we generated four types of synthetic graphs: BA or power-law graphs, ER graphs, LFR community graphs, and a mixture of different grid/lattice graphs. Figure \ref{fig:graph_types} illustrates the structure of these graph types.

For the fourth task, we use community based 20 LFR graphs to estimate top-k (k = 20, 50, \& 100) high ranking nodes.
\subsection{Graph Sampling}
As discussed in Section \ref{sec:est_node_edge}, random walks are the primary way to sample from graphs. In real-world applications, simple random walks are more practical, while MH-walks (\texttt{mh}) or max-degree walks are difficult to implement. Similarly, in the research literature, a \textit{burn-in} (where a fixed number of initial nodes are dropped from the walk) is often used to lose dependence on the initial seed node; however, in practical situations, querying nodes can have an associated cost, and discarding them can be wasteful. 

For our experiments, and for a fair comparison with baseline methods, we use both \texttt{MH} walks and \texttt{srw} walks with \textit{burn-in} for generation of LLM prompts. For the task of estimating the number of clusters and graph structure, we use more practical \texttt{srw} walk to sample the graphs.

\subsection{Prompt Generation}
Previous works on benchmarking LLMs on graph-algorithmic tasks generate input prompts for LLMs by encoding the entire graph structure into natural language in the form of an edgelist or adjacency list. However, this approach fails with large graphs due to the context length limitations.

\subsubsection*{Prompting Strategy}

We assume that random walk sampling in a graph provides us with the name of the node, the degree of the node, and the ability to continue the walk by accessing neighbors of the node. Based on the information derived from the walks, we create walk statistics that are relevant to the estimation task. This approach helps limit the number of tokens required in the prompt, enabling us to scale the evaluations to very large graphs. 

\begin{figure}
    \centering
    \includegraphics[width=0.97\linewidth]{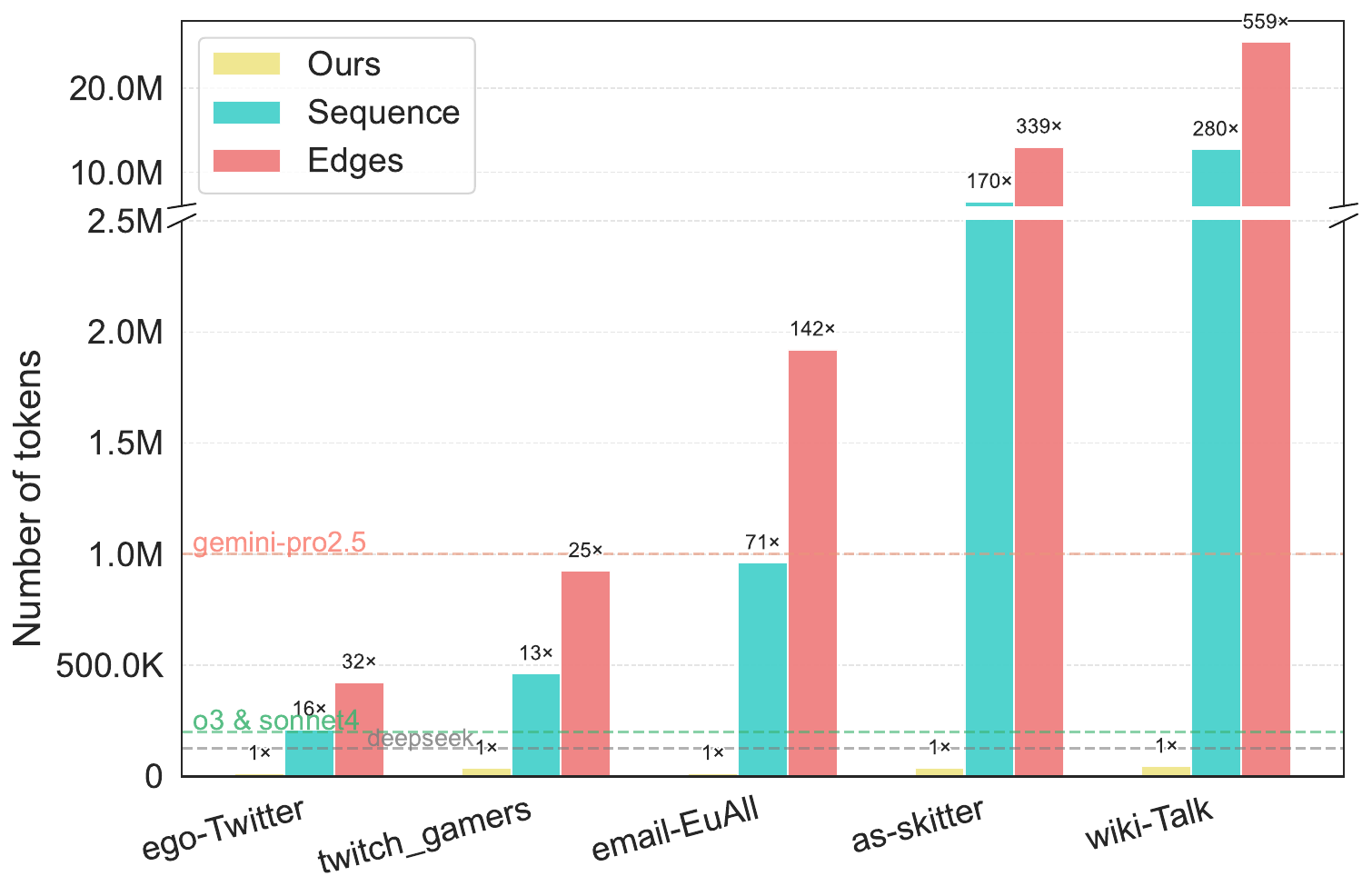}
    \caption{\textcolor{black}{Figure shows comparison of number of tokens (with tiktoken) in our statistics-based prompt vs naive encoding of real-world graphs. Horizontal dashed line indicates the context length of LLMs.}}

    \label{fig:plot_num_tokens}
\end{figure}

Based on their prior knowledge about graph theory and the walk statistics in the prompt, LLMs can reason and estimate the required properties of the graphs. Figures \ref{fig:prompt_template}-\ref{fig:prompt_template_topk} provide the prompt templates for all tasks. For more details on the datasets and prompt construction, please refer to Appendix \ref{sec:appendix_experiment_parameters}. \textcolor{black}{Figure \ref{fig:plot_num_tokens} shows a comparison of the number of tokens in prompts for real-world graph datasets (graph size estimation task) generated using our approach, using walk sequences in the prompt, and using edgelist to encode graphs in previous works. With our approach, the number of tokens are significantly smaller and well within LLMs' context lengths. }

\subsection{Results}

\begin{table*}[ht]
\centering

\resizebox{0.85\linewidth}{!}{%
\begin{tabular}{c lccc ccc ccc c}
\toprule
\textbf{Graph} &  & \multicolumn{3}{c}{Small} & \multicolumn{3}{c}{Medium} & \multicolumn{3}{c}{Large} & Rank \\
\textbf{Type} & Method & Median & Mean & Std & Median & Mean & Std & Median & Mean & Std &  \\
\midrule
\multirow{10}{*}{\rotatebox[origin=c]{90}{BA}} 
 & uniform\textsuperscript{\textcolor{carnelian}{\dag}}      & \textbf{13.63} & 13.13 & 6.77 &  \textbf{4.20} &  5.96 & 4.52 &  \textbf{0.60} &  0.83 & 0.78 & 1 \\
 & mh\textsuperscript{\textcolor{carnelian}{\dag}}           & 14.28 & 17.89 &10.28 & 10.37 & 10.14 & 5.79 & 12.17 & 11.70 & 2.72 & 2 \\
 & max\_degree\textsuperscript{\textcolor{carnelian}{\dag}}  & 25.83 & 27.03 & 9.77 & 27.43 & 27.07 & 3.01 & 27.21 & 27.01 & 0.96 & 8 \\
 & return\_walk\textsuperscript{\textcolor{greenncs}{\dag}}       & 48.10 & 75.06 &100.33& 20.02 & 38.22 &33.76 & 22.79 & 26.31 &17.12 &11 \\
 \noalign{\vskip 1.3mm}
 & \textcolor{lightcoral}{gemini-2.5-pro (mh)}\textsuperscript{\textcolor{carnelian}{\dag}}   & 20.44 & 32.09 &30.61 & 40.19 & 52.32 &46.43 & 20.03 & 43.62 &50.18 &9 \\
 & \textcolor{lightcoral}{o3 (mh)}\textsuperscript{\textcolor{carnelian}{\dag}}   & 17.60 & 50.31 &86.82 &  \textcolor{greenmun}{9.63} & 37.64 &87.75 & \textcolor{greenmun}{13.08} & 22.17 &30.76 &3 \\
 & \textcolor{lightcoral}{sonnet-4 (mh)}\textsuperscript{\textcolor{carnelian}{\dag}}   & 23.09 & 35.95 &43.15 & 12.26 & 18.79 &18.96 & 30.75 & 32.45 &22.76 &5 \\
  & \textcolor{lightcoral}{deepseek-v3.1 (mh)}\textsuperscript{\textcolor{carnelian}{\dag}}   & 21.95 & 23.99 &14.54 & 12.11 & 18.40 &17.59 & 14.01 & 41.51 &72.55 &4 \\
 \noalign{\vskip 1.2mm}
 & \textcolor{denim}{gemini-2.5-pro (srw)}\textsuperscript{\textcolor{greenncs}{\dag}}  & 23.32 & 43.65 &46.38 & 62.02 & 60.05 &44.42 & 52.56 & 53.74 &28.78 &12 \\
 & \textcolor{denim}{o3 (srw)}\textsuperscript{\textcolor{greenncs}{\dag}}  & \textcolor{greenmun}{16.01} & 42.41 &62.15 & 30.58 & 37.84 &30.07 & 25.47 & 24.04 & 9.65 &6 \\
 & \textcolor{denim}{sonnet-4 (srw)}\textsuperscript{\textcolor{greenncs}{\dag}}  & 20.48 & 30.65 &29.21 & 32.12 &151.18 &464.27& 36.52 & 36.66 &17.13 &10 \\
& \textcolor{denim}{deepseek-v3.1 (srw)}\textsuperscript{\textcolor{greenncs}{\dag}}   & 20.23 & 22.09 &9.97 & 26.87 & 36.13 &26.30 & 26.97 & 35.38 &15.55 &7 \\

\midrule
\multirow{10}{*}{\rotatebox[origin=c]{90}{ER}} 
 & uniform\textsuperscript{\textcolor{carnelian}{\dag}}      & 12.68 & 14.32 & 7.07 &  \textbf{4.40} &  6.24 & 3.89 &  \textbf{0.77} &  0.92 & 0.74 & 1 \\
 & mh\textsuperscript{\textcolor{carnelian}{\dag}}           & 13.54 & 15.86 &10.80 &  6.30 &  6.11 & 4.44 &  2.39 &  2.44 & 1.09 & 4 \\
 & max\_degree\textsuperscript{\textcolor{carnelian}{\dag}}  & 15.03 & 17.95 &10.22 &  6.48 &  6.04 & 3.69 &  4.80 &  5.49 & 2.19 & 5 \\
 & return\_walk\textsuperscript{\textcolor{greenncs}{\dag}}       & 45.86 & 75.96 &82.72 & 62.29 &118.98 &128.25& 54.87 & 68.69 &58.08 &12 \\
  \noalign{\vskip 1.3mm}
 & \textcolor{lightcoral}{gemini-2.5-pro (mh)}\textsuperscript{\textcolor{carnelian}{\dag}}   & 20.55 & 27.22 &25.72 & 16.97 & 26.89 &19.43 & 12.12 & 22.92 &28.12 &8 \\
 & \textcolor{lightcoral}{o3 (mh)}\textsuperscript{\textcolor{carnelian}{\dag}}   &  \textcolor{greenmun}{\textbf{7.59}} & 10.04 & 7.01 &  8.90 &  7.72 & 5.27 &  \textcolor{greenmun}{3.41} &  9.37 &12.75 &3 \\
 & \textcolor{lightcoral}{sonnet-4 (mh)}\textsuperscript{\textcolor{carnelian}{\dag}}   & 23.70 & 68.05 &158.00& 48.08 & 38.01 &22.04 & 36.83 & 32.30 &23.51 &11 \\
& \textcolor{lightcoral}{deepseek-v3.1 (mh)}\textsuperscript{\textcolor{carnelian}{\dag}}   & 9.41 & 16.43 &16.88 & \textcolor{greenmun}{4.50} & 13.88 &21.02 & 4.04 & 10.89 &14.06 &2 \\
  \noalign{\vskip 1.2mm}
 & \textcolor{denim}{gemini-2.5-pro (srw)}\textsuperscript{\textcolor{greenncs}{\dag}}  & 33.87 & 34.50 &28.61 &  9.40 & 14.21 &14.56 &  8.08 & 15.28 &14.08 &9 \\
 & \textcolor{denim}{o3 (srw)}\textsuperscript{\textcolor{greenncs}{\dag}}  &  9.59 & 16.98 &21.02 & 12.95 & 15.79 &12.36 &  5.57 &  8.74 & 9.27 &6 \\
 & \textcolor{denim}{sonnet-4 (srw)}\textsuperscript{\textcolor{greenncs}{\dag}}  & 13.98 & 21.46 &17.28 & 25.23 & 29.93 &24.55 & 19.00 & 25.87 &20.78 &10 \\
& \textcolor{denim}{deepseek-v3.1 (srw)}\textsuperscript{\textcolor{greenncs}{\dag}}   & 18.41 & 20.35 &11.57 & 6.00 & 10.40 &11.76 & 6.87 & 18.78 &22.76 &7 \\
\midrule
\multirow{10}{*}{\rotatebox[origin=c]{90}{GRP}} 
 & uniform\textsuperscript{\textcolor{carnelian}{\dag}}      & 14.07 & 14.61 & 6.27 &  4.42 &  6.22 & 3.93 &  \textbf{0.56} &  0.76 & 0.74 & 4 \\
 & mh\textsuperscript{\textcolor{carnelian}{\dag}}           &  7.21 & 10.04 & 8.28 &  4.04 &  6.06 & 5.93 &  2.51 &  2.70 & 1.29 & 1 \\
 & max\_degree\textsuperscript{\textcolor{carnelian}{\dag}}  & 19.41 & 21.46 &13.95 &  5.48 &  6.53 & 5.50 &  5.64 &  5.63 & 1.66 & 7 \\
 & return\_walk\textsuperscript{\textcolor{greenncs}{\dag}}       & 82.10 & 83.87 &57.52 & 65.99 &102.77 &96.60 & 21.41 & 55.97 &88.00 &12 \\
  \noalign{\vskip 1.3mm}
 & \textcolor{lightcoral}{gemini-2.5-pro (mh)}\textsuperscript{\textcolor{carnelian}{\dag}}   & 36.64 & 81.83 &170.06& 15.37 & 25.42 &19.96 &  5.38 &  8.53 & 8.61 &10 \\
 & \textcolor{lightcoral}{o3 (mh)}\textsuperscript{\textcolor{carnelian}{\dag}}   & 11.90 & 25.81 &24.44 &  \textcolor{greenmun}{\textbf{3.09}} &  7.83 &10.35 &  \textcolor{greenmun}{2.81} &  7.87 &14.28 &2 \\
 & \textcolor{lightcoral}{sonnet-4 (mh)}\textsuperscript{\textcolor{carnelian}{\dag}}   & 19.53 & 28.50 &30.06 & 11.20 & 25.59 &23.52 & 17.54 & 25.82 &24.59 &9 \\
   & \textcolor{lightcoral}{deepseek-v3.1 (mh)}\textsuperscript{\textcolor{carnelian}{\dag}}   & 15.25 & 19.63 &14.95 & 3.52 & 9.37 &16.00 & 3.06 & 10.13 &15.75 &5 \\
  \noalign{\vskip 1.2mm}
 & \textcolor{denim}{gemini-2.5-pro (srw)}\textsuperscript{\textcolor{greenncs}{\dag}}  & 22.48 & 26.24 &30.81 &  7.98 & 13.55 &12.43 & 16.84 & 19.89 &20.01 &8 \\
 & \textcolor{denim}{o3 (srw)}\textsuperscript{\textcolor{greenncs}{\dag}}  &  \textcolor{greenmun}{\textbf{6.73}} & 11.00 & 9.30 &  6.81 &  7.13 & 6.55 &  4.94 &  6.36 & 4.25 &3 \\
 & \textcolor{denim}{sonnet-4 (srw)}\textsuperscript{\textcolor{greenncs}{\dag}}  & 23.70 & 26.66 &20.79 & 12.50 & 21.32 &20.28 & 44.16 & 40.62 &22.01 &11 \\
& \textcolor{denim}{deepseek-v3.1 (srw)}\textsuperscript{\textcolor{greenncs}{\dag}}   & 11.83 & 19.54 &18.34 & 6.40 & 8.68 &12.24 & 4.94 & 8.84 &9.55 &6 \\
\bottomrule
\end{tabular}
}
\caption{Comparison of relative error percentages for node estimations. \textbf{Bold} values show best performance among all methods and \textcolor{greenmun}{green} values show best performance among LLMs. Ranking is based on median values. \textsuperscript{\textcolor{carnelian}{\dag}} methods require unbiased sampling (difficult in real-world applications); \textsuperscript{\textcolor{greenncs}{\dag}} methods require \texttt{srw} walks.}
\vspace{-4mm}

\label{tab:node_est}
\end{table*}

\begin{table*}[ht]
\centering
\resizebox{0.97\linewidth}{!}{%
\begin{tabular}{l ccc ccc ccc ccc ccc c}
\toprule
 & \multicolumn{3}{c}{ego-Twitter} & \multicolumn{3}{c}{twitch-gamers} & \multicolumn{3}{c}{email-EuAll} & \multicolumn{3}{c}{as-skitter} & \multicolumn{3}{c}{wiki-Talk} & Rank \\
\textbf{Method} & Median & Mean & Std & Median & Mean & Std & Median & Mean & Std & Median & Mean & Std & Median & Mean & Std &  \\
\midrule
uniform\textsuperscript{\textcolor{carnelian}{\dag}} & \textbf{0.83} & 1.06 & 0.81 &  \textbf{0.52} &  0.69 & 0.53 &  \textbf{0.68} &  0.80 & 0.66 & \textbf{0.17} & 0.21 & 0.16 & \textbf{0.10} & 0.18 & 0.14 &  1\\
mh\textsuperscript{\textcolor{carnelian}{\dag}} & 51.02 & 52.04 & 3.84 &  59.62 &  59.77 & 1.10 &  136.20 &  134.75 & 5.87 & 75.21 & 75.40 & 1.27 & 181.04 & 181.16 & 0.41 &  10\\
  \noalign{\vskip 1.3mm}
\textcolor{lightcoral}{gemini-2.5-pro (mh)}\textsuperscript{\textcolor{carnelian}{\dag}}  & 66.04 & 119.25 & 117.18 & \textcolor{greenmun}{36.64} & 31.29 & 7.52 & \textcolor{greenmun}{19.06} & 23.03 & 16.44 & \textcolor{greenmun}{30.01} & 49.82 & 50.01 & 64.37 & 227.64 & 371.45 & 4 \\
\textcolor{lightcoral}{o3 (mh)}\textsuperscript{\textcolor{carnelian}{\dag}}              & 33.60 & 174.23 & 314.38 & 36.65 & 38.38 & 4.65 & 57.48 & 58.00 & 4.01 & 40.99 & 37.33 & 8.03 & 58.14 & 59.65 & 3.56 & 6 \\
\textcolor{lightcoral}{sonnet-4 (mh)}\textsuperscript{\textcolor{carnelian}{\dag}}        & 33.58 & 37.77 & 9.82 & 40.52 & 42.60 & 18.73 & 62.19 & 61.16 & 7.09 & 49.84 & 44.24 & 23.93 & 64.42 & 61.24 & 8.84 & 8 \\
\textcolor{lightcoral}{deepseek-v3.1 (mh)}\textsuperscript{\textcolor{carnelian}{\dag}}       & \textcolor{greenmun}{33.40} & 36.92 & 7.76 & \textcolor{greenmun}{36.64} & 36.64 & 0.0 & 57.48 & 52.25 & 9.33 & 42.32 & 52.17 & 13.48 & 64.37 & 64.37 & 0.01 & 7 \\

\noalign{\vskip 1.2mm}
\textcolor{denim}{gemini-2.5-pro (srw)}\textsuperscript{\textcolor{greenncs}{\dag}}      & 44.65 & 205.33 & 320.05 & 52.41 & 67.20 & 37.00 & 29.98 & 70.84 & 65.61 & 32.39 & 55.44 & 59.27 & \textcolor{greenmun}{27.65} & 98.27 & 111.25 & 2 \\
\textcolor{denim}{o3 (srw)}\textsuperscript{\textcolor{greenncs}{\dag}}                  & 51.85 & 268.31 & 481.66 & 52.41 & 52.18 & 0.53 & 28.84 & 41.32 & 51.30 & 49.84 & 45.77 & 9.27 & 33.03 & 32.37 & 1.46 & 3 \\
\textcolor{denim}{sonnet-4 (srw)}\textsuperscript{\textcolor{greenncs}{\dag}}            & 63.10 & 59.66 & 5.73 & 55.39 & 50.48 & 18.45 & 46.63 & 46.63 & 13.34 & 50.43 & 54.17 & 7.56 & 53.95 & 53.37 & 9.86 & 9 \\
\textcolor{denim}{deepseek-v3.1 (srw)}\textsuperscript{\textcolor{greenncs}{\dag}}            & 51.83 & 43.50 & 18.61 & 52.41 & 53.60 & 2.66 & 29.99 & 30.71 & 1.52 & 50.21 & 50.21 & 0.00 & 34.38 & 34.38 & 0.00 & 5 \\
\bottomrule
\end{tabular}}

\caption{Comparison of relative error percentages for node estimations for real-world datasets. Ranking is based on median values. max\_degree and return\_walk are omitted due to large execution time. \textsuperscript{\textcolor{carnelian}{\dag}} indicate methods that are difficult to implement in real-world setting, while \textsuperscript{\textcolor{greenncs}{\dag}} methods can be used in real-world tasks.}
\vspace{-5mm}

\label{tab:real_world_node_est}
\end{table*}

\subsubsection{Graph size estimation}

Table \ref{tab:node_est} \& \ref{tab:real_world_node_est} show the comparison of relative error percentages in the estimation of the number of nodes. Edge estimation results are included in Appendix \ref{sec:appendix}.

\textit{Observations}
\vspace{-3mm}

\begin{itemize}
    \item Conventional methods \texttt{uniform sampling} and \texttt{mh}, perform well with lower error values, while \texttt{max\_degree} and \texttt{return\_walk} have higher error values. However, these methods also require unbiased sampling of graphs, which is difficult in real-world applications.   

    \item For synthetic datasets, LLMs can provide reasonable estimates of graph size with median relative error percentages below 20\% in many cases. Overall, o3 performs best among all LLMs, sometimes even better than the baseline methods. For real-world much larger datasets, error percentages are higher with gemini-2.5-pro providing the best estimations, however they still outperform baseline method like \texttt{mh}.

    \item For baseline methods, mean and median error values do not differ much, implying consistent estimations over multiple graphs. However, LLMs show larger differences, suggesting higher variation in the estimated values. For a few predictions, LLMs significantly overestimate the number of nodes and edges, leading to skewed mean error values.

    \item Prompts for LLMs were created with two walk settings: \texttt{mh} and \texttt{srw}. \texttt{srw} is more practical to implement, its overall performance is lower with 78\% (BA), 51\% (GRP) and 9\% (real-world) relative to LLMs with \texttt{mh} setting. Results on the estimation of edges follow similar patterns. 

\end{itemize}

\subsubsection{Estimate Number of Communities}
\begin{figure}
    \centering
    \includegraphics[width=0.85\linewidth]{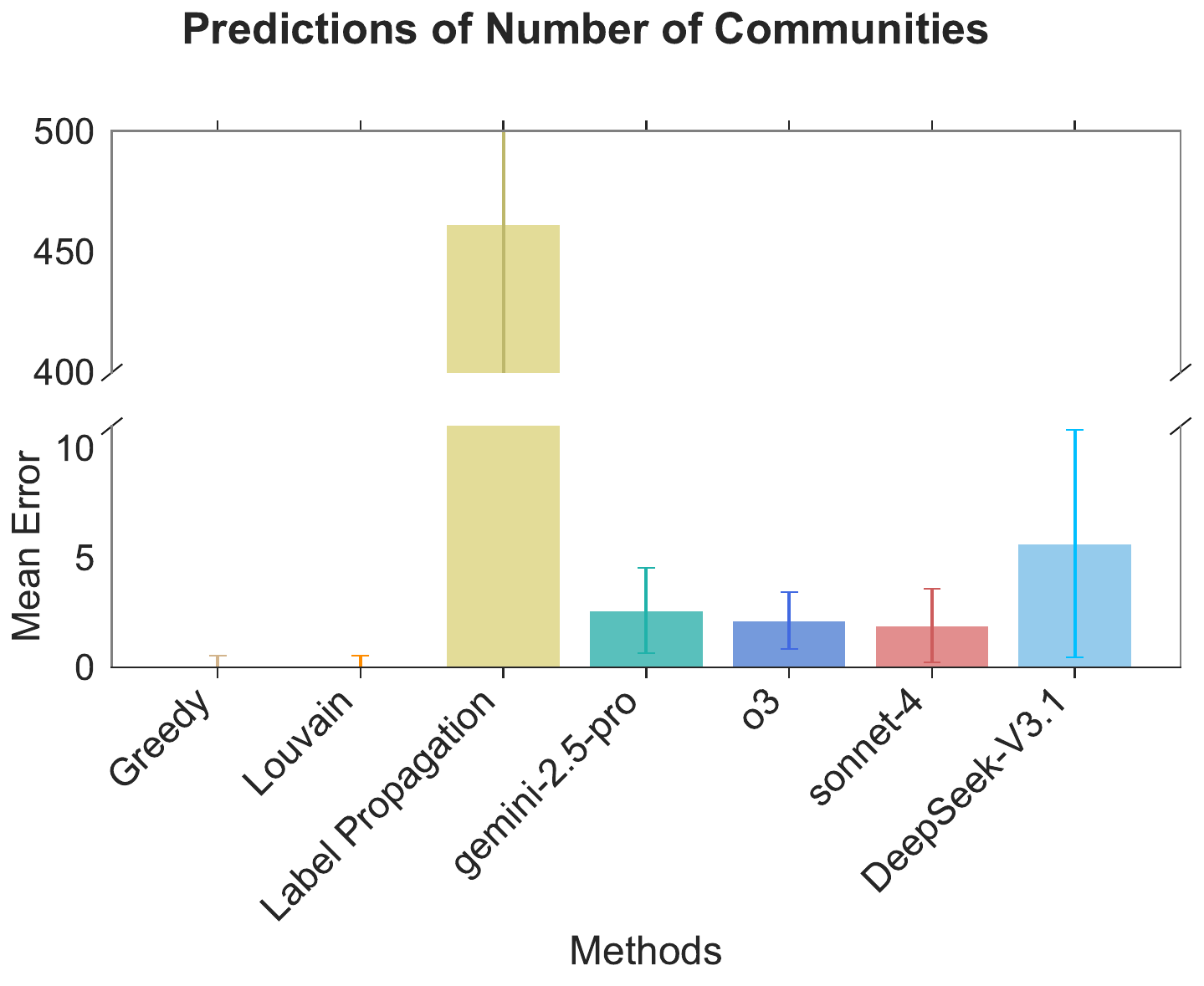}
    \caption{Figure shows the mean error in prediction of number of communities present in the LFR graphs.}
    \label{fig:cluster_pred}
\end{figure}

Figure \ref{fig:cluster_pred} shows the mean absolute error in predicting the number of communities in the LFR graphs.

\textit{Observations}
\vspace{-3mm}
\begin{itemize}
    \item Among baseline methods, \texttt{Greedy} and \texttt{Louvain} are able to detect communities accurately with almost no error with subgraphs created based on random walks, while \texttt{Label propagation} fails with significant errors.

    \item LLMs detected communities with error rates ranging from 1.9 to 2.6 with communities size ranging from 5 to 12 in numbers.
\end{itemize}

\subsubsection{Estimate Graph Structure}
Table \ref{tab:struct_pred} shows the accuracy of the prediction of the graph structure by LLMs. 

\textit{Observations}
\vspace{-3mm}
\begin{itemize}
    \item All LLMs were able to easily identify the grid structure due to uniform degree distribution.

    \item Overall, the o3 model performed best, with low accuracy only in the LFR graphs.

    \item For BA and LFR graphs, the o3 and gemini-2.5-pro models showed a contrast in their predictions. While o3's predictions were biased towards BA, gemini model's predictions were biased towards LFR. However, neither model predicted the label "ER" for any BA/LFR graphs, demonstrating their ability to distinguish between the uniform degree distribution of ER graphs and the skewed degree distribution of BA and LFR graphs \textcolor{black}{(Figure \ref{fig:degree_distribution})}. 
\end{itemize}

\begin{table}[h!]
\centering
 \resizebox{0.90\linewidth}{!}{%
\begin{tabular}{lcccc}
\hline
\textbf{Models} & \textbf{BA} & \textbf{ER} & \textbf{LFR} & \textbf{Grid} \\
\hline
gemini-2.5-pro   & 33.3\% & 73.3\% & 80.0\% & 100.0\% \\
o3   & 93.3\% & 93.3\% & 26.7\% & 100.0\% \\
sonnet-4   & 100.0\% & 13.3\% & 6.7\%  & 100.0\% \\
DeepSeek-V3.1   & 80.0\% & 66.67\% & 66.67\% & 100.0\% \\
\hline
\end{tabular}
}
\caption{Accuracy in graph structure prediction by LLMs}
\vspace{-1mm}

\label{tab:struct_pred}
\end{table}

\begin{table}[h!]
\centering

 \resizebox{\linewidth}{!}{%
\begin{tabular}{lrrr}
\toprule
\textbf{Model} & \textbf{Betweenness} & \textbf{Closeness} & \textbf{PageRank} \\
\midrule
gemini-2.5-pro   & $6.50 \pm 7.43$   & $9.25 \pm 8.41$   & $27.50 \pm 18.41$ \\
o3       & $31.50 \pm 14.20$ & $35.00 \pm 11.73$ & $81.00 \pm 19.85$ \\
sonnet   & $15.25 \pm 10.06$ & $23.75 \pm 16.11$ & $42.75 \pm 28.39$ \\
DeepSeek-V3.1 & $23.00 \pm 13.64$ & $20.00 \pm 16.43$ & $28.50 \pm 23.03$ \\
\bottomrule
\end{tabular}
}

\caption{\textcolor{black}{{Precision@20} (\%) for node centrality values.}}
\vspace{-1mm}
\label{tab:node_ranking}
\end{table}

\subsubsection{\textcolor{black}{Estimate Influential Nodes in the Graph}}
\textcolor{black}{Table \ref{tab:node_ranking} shows the prediction accuracy for node ranking by LLMs.}

\textit{\textcolor{black}{Observations}}
\vspace{-2mm}

\begin{itemize}
    \item \textcolor{black}{Betweenness and Closeness ranking measures are based on shortest-path calculations and are therefore difficult to estimate with simple random walks. However, the o3 model can achieve a mean 30–35\% precision@20, based on the degree distribution of nodes in the walks, while other models perform poorly.}

    \item \textcolor{black}{PageRank values are often correlated with how frequently a simple random walk would visit each node. Therefore, all models perform better on this ranking measure. However, the o3 model significantly outperforms other LLMs in its predictions.}
\end{itemize}

\section{Discussion}
\label{sec:discussion}

\subsection*{\textcolor{black}{Large Graph Analysis with LLMs}}
\textcolor{black}{In our experiments, we observe that LLMs can leverage random-walk statistics and the degree distribution of nodes to understand graph structure and estimate many properties of a given graph. Without any explicit instructions in the prompt, models use reasoning abilities to distinguish between different types of graphs, identify community patterns in the walks, influential nodes in the graphs ,and produce graph-size estimates.}

\subsection*{\textcolor{black}{Practical recommendations for Large Graph Analysis with LLMs}}
\textcolor{black}{The following points can be used when devising large graph analysis with LLMs:}

\begin{enumerate}[label=(\roman*), itemsep=-2mm]
\color{black}
    \item LLMs can provide reasonable graph size estimates from walk statistics (relative error below 20\% in many cases) . While unbiased sampling (mh-walk) is ideal in principle, it can be difficult to implement in practical settings, and a simple random walk can also provide good estimates for graph size.

    \item Using multiple runs for analysis and using median or robust aggregation of estimates helps to identify any potential outliers estimations.

    \item Graph size estimates generally improve as the random-walk budget increases (Figure \ref{fig:node_est_change}).

    \item For community detection, longer walks can help to identify communities more easily, with repeated visits within a region providing stronger evidence of clustering. This process can also be split into multiple prompts with different seed nodes or API entry endpoints.

    \item LLMs use degree distributions to detect graph structure. Using multiple short-to-medium walks from diverse start nodes can help reduce local-region bias.

    \item Estimating PageRank-like importance is more feasible than shortest-path based ranking measures. However, model choice plays an important role with estimation accuracy.

\end{enumerate}

\section{Conclusion}

\textcolor{black}{In this work, we propose a new benchmark \texttt{EstGraph} and evaluate the abilities of reasoning-based LLMs on estimation tasks for large graphs that have not previously been studied using LLM reasoning. Our experiments show with appropriate prompting method, LLMs can be used to estimate properties of large graphs with reasonable accuracy.}

\section*{Limitations}
In our work, we have evaluated the capabilities of reasoning LLMs on the task of estimating properties of large graphs. This topic has been extensively researched with algorithmic-based estimation approaches under various settings. Although we have included a wide variety of tasks with various types of graph structures, however, our evaluation still does not cover the entire range of possible tasks. 

Another potential limitation is the lack of ablation over multiple hyperparameters like sampling size, burn-in rate, random walk initializations, and so on. Due to the high cost and token consumption of reasoning LLM APIs, it was not feasible to run experiments under many different hyperparameter settings. As models continue to improve and the cost per token decreases, it may become possible to extend these experiments to cover more comprehensive settings.

\section*{Potential Risks Discussion}
Our work focuses on the evaluation of LLMs to estimate graph properties from random walk statistics. The datasets used in this study are either synthetic or already publicly available. Given the current scope, our work does not pose any potential risks.

\section*{Acknowledgments}

This work was partially supported by JSPS Grant-in-Aid for Scientific Research (grant number 25K03231, 23H03451). This paper is based on results obtained from the BRIDGE Program (R7-H05), implemented by the Cabinet Office, Government of Japan.

\bibliography{latex/references}

\appendix

\section{Supplementary Results}
\label{sec:appendix}

\subsection{Graph Size Estimation}
Tables \ref{tab:edge_est} and \ref{tab:real_data_edge_est} show the relative error percentages for edge estimations on synthetic and real-world datasets. For synthetic datasets, the performance of LLMs for ER and GRP graphs is very similar to baseline methods such as \texttt{mh}. For BA dataset, they show higher error values, potentially due to the skewed degree distribution of the graphs. For real-world datasets, most results follow a similar pattern as node estimation.

\begin{table*}[ht]
\centering

\resizebox{0.9\linewidth}{!}{%
\begin{tabular}{c lccc ccc ccc c}
\toprule
\textbf{Graph} &  & \multicolumn{3}{c}{Small} & \multicolumn{3}{c}{Medium} & \multicolumn{3}{c}{Large} & Rank \\
\textbf{Type} & Method & Median & Mean & Std & Median & Mean & Std & Median & Mean & Std &  \\
\midrule
\multirow{10}{*}{\rotatebox[origin=c]{90}{BA}} 
 & uniform\textsuperscript{\textcolor{carnelian}{\dag}}      &  11.15 & 11.69 &  7.84 &  \textbf{8.22} &  8.16 &  5.16 &  \textbf{0.87} &  1.57 &   1.72 & 1   \\
 & mh\textsuperscript{\textcolor{carnelian}{\dag}}           & 15.24 & 23.03 & 25.41 & 12.14 & 14.35 &  8.32 & 13.25 & 12.77 &   2.47 & 2 \\
 & max\_degree\textsuperscript{\textcolor{carnelian}{\dag}}  & 16.99 & 18.36 & 12.78 & 22.41 & 21.45 &  4.89 & 22.44 & 22.71 &   1.80 & 4 \\
 & return\_walk\textsuperscript{\textcolor{greenncs}{\dag}}       & 34.14 & 53.57 & 70.90 & 39.04 & 38.63 & 25.64 & 18.73 & 23.90 &  16.94 & 12 \\
   \noalign{\vskip 1.3mm}
 & \textcolor{lightcoral}{gemini-2.5-pro (mh)}\textsuperscript{\textcolor{carnelian}{\dag}}   & 28.92 & 29.43 & 23.11 & 18.88 & 23.00 & 20.40 & 24.53 & 29.50 &  22.12 & 6 \\
 & \textcolor{lightcoral}{o3 (mh)}\textsuperscript{\textcolor{carnelian}{\dag}}   & 37.50 & 61.04 & 71.32 & \textcolor{greenmun}{18.37} & 39.76 & 70.50 & 29.58 & 26.76 &  21.31 & 7 \\
 & \textcolor{lightcoral}{sonnet-4 (mh)}\textsuperscript{\textcolor{carnelian}{\dag}}   & 24.63 & 46.82 & 58.71 & 22.62 & 24.55 & 16.84 & \textcolor{greenmun}{20.35} & 24.27 &  19.22 & 5 \\
 & \textcolor{lightcoral}{deepseek-v3.1 (mh)}\textsuperscript{\textcolor{carnelian}{\dag}}   & 35.71 & 38.86 & 21.19 & 23.63 & 29.85 & 25.80 & 26.87 & 39.40 &  34.76 & 8 \\

   \noalign{\vskip 1.2mm}
 & \textcolor{denim}{gemini-2.5-pro (srw)}\textsuperscript{\textcolor{greenncs}{\dag}}  & 23.78 & 32.31 & 24.53 & 29.08 & 39.39 & 22.72 & 33.61 &117.26 & 309.52 & 9 \\
 & \textcolor{denim}{o3 (srw)}\textsuperscript{\textcolor{greenncs}{\dag}}  & 48.63 & 74.69 & 62.30 & 18.65 & 40.39 & 39.84 & 22.30 & 37.83 &  40.37 & 11 \\
 & \textcolor{denim}{sonnet-4 (srw)}\textsuperscript{\textcolor{greenncs}{\dag}}  &  \textcolor{greenmun}{\textbf{8.59}} & 49.75 & 86.99 & 19.30 &223.78 &765.48 & 30.38 & 42.19 &  35.36 & 3 \\
   & \textcolor{denim}{deepseek-v3.1 (srw)}\textsuperscript{\textcolor{greenncs}{\dag}}   & 31.76 & 31.37 & 19.04 & 25.69 & 48.00 & 63.83 & 30.41 & 94.17 &  183.97 & 10 \\

\midrule
\multirow{10}{*}{\rotatebox[origin=c]{90}{ER}} 
 & uniform\textsuperscript{\textcolor{carnelian}{\dag}}      & 14.20 & 14.65 &  7.06 &  3.94 &  6.08 &  3.93 &  \textbf{0.70} &  0.89 &   0.78 & 2 \\
 & mh\textsuperscript{\textcolor{carnelian}{\dag}}           & 15.38 & 16.11 & 13.14 &  2.33 &  4.95 &  4.23 &  1.23 &  1.59 &   1.14 & 3 \\
 & max\_degree\textsuperscript{\textcolor{carnelian}{\dag}}  & 11.40 & 14.10 & 10.48 &  \textbf{2.29} &  3.04 &  2.38 &  1.50 &  1.84 &   1.22 & 1 \\
 & return\_walk\textsuperscript{\textcolor{greenncs}{\dag}}       & 38.52 & 56.93 & 72.63 & 33.92 & 50.31 & 38.22 & 44.33 & 63.13 &  44.35 &12 \\
   \noalign{\vskip 1.3mm}
 & \textcolor{lightcoral}{gemini-2.5-pro (mh)}\textsuperscript{\textcolor{carnelian}{\dag}}   & 13.26 & 17.08 & 14.10 &  \textcolor{greenmun}{6.94} & 10.16 & 10.35 &  4.07 & 19.42 &  23.64 & 4 \\
 & \textcolor{lightcoral}{o3 (mh)}\textsuperscript{\textcolor{carnelian}{\dag}}   & 11.77 & 13.02 &  7.08 &  7.46 &  8.18 &  5.94 &  5.81 & 10.81 &  12.43 & 5 \\
 & \textcolor{lightcoral}{sonnet-4 (mh)}\textsuperscript{\textcolor{carnelian}{\dag}}   & 23.21 & 70.59 &165.41 & 46.35 & 37.50 & 22.32 & 34.30 & 33.10 &  21.22 & 11 \\
 & \textcolor{lightcoral}{deepseek-v3.1 (mh)}\textsuperscript{\textcolor{carnelian}{\dag}}   & 15.62 & 19.08 & 19.72 & 6.95 & 17.92 & 24.69 & 21.30 & 26.62 &  24.26 & 9 \\

   \noalign{\vskip 1.2mm}
 & \textcolor{denim}{gemini-2.5-pro (srw)}\textsuperscript{\textcolor{greenncs}{\dag}}  & 11.63 & 19.78 & 17.91 & 13.23 & 16.93 & 14.65 &  \textcolor{greenmun}{2.39} &  4.30 &   5.63 & 6 \\
 & \textcolor{denim}{o3 (srw)}\textsuperscript{\textcolor{greenncs}{\dag}}  &  \textcolor{greenmun}{\textbf{8.67}} & 14.71 & 22.24 & 17.23 & 13.97 & 10.25 &  3.71 &  7.42 &   9.48 & 7 \\
 & \textcolor{denim}{sonnet-4 (srw)}\textsuperscript{\textcolor{greenncs}{\dag}}  & 17.95 & 21.61 & 16.36 & 23.38 & 28.57 & 23.82 & 22.46 & 25.06 &  25.38 & 10 \\
   & \textcolor{denim}{deepseek-v3.1 (srw)}\textsuperscript{\textcolor{greenncs}{\dag}}   & 23.05 & 27.36 & 21.01 & 10.88 & 14.89 & 14.14 & 5.64 & 25.08 &  28.62 & 8 \\

\midrule
\multirow{10}{*}{\rotatebox[origin=c]{90}{GRP}} 
 & uniform\textsuperscript{\textcolor{carnelian}{\dag}}      & 14.28 & 14.07 &  6.61 &  \textbf{4.53} &  6.05 &  3.84 &  \textbf{0.53} &  0.80 &   0.83 & 2 \\
 & mh\textsuperscript{\textcolor{carnelian}{\dag}}           &  \textbf{8.17} & 10.31 &  7.97 &  5.46 &  6.29 &  5.17 &  1.80 &  1.74 &   1.04 & 1 \\
 & max\_degree\textsuperscript{\textcolor{carnelian}{\dag}}  & 13.38 & 20.93 & 16.61 &  6.47 &  7.03 &  4.80 &  1.55 &  1.58 &   1.27 & 4 \\
 & return\_walk\textsuperscript{\textcolor{greenncs}{\dag}}       & 47.63 & 60.42 & 50.86 & 31.29 & 85.20 &112.12 & 62.51 & 71.91 &  56.81 &12 \\
   \noalign{\vskip 1.3mm}
 & \textcolor{lightcoral}{gemini-2.5-pro (mh)}\textsuperscript{\textcolor{carnelian}{\dag}}   & 29.72 & 76.25 &188.73 &  8.21 & 12.89 & 13.31 &  4.17 &  6.89 &   8.34 & 9 \\
 & \textcolor{lightcoral}{o3 (mh)}\textsuperscript{\textcolor{carnelian}{\dag}}   & 22.84 & 22.95 & 22.95 &  6.65 & 11.23 &  9.88 &  4.41 & 10.50 &  14.76 & 7 \\
 & \textcolor{lightcoral}{sonnet-4 (mh)}\textsuperscript{\textcolor{carnelian}{\dag}}   & 25.04 & 32.57 & 30.10 & 18.53 & 26.29 & 22.24 & 13.91 & 26.59 &  25.41 & 10 \\
 & \textcolor{lightcoral}{deepseek-v3.1 (mh)}\textsuperscript{\textcolor{carnelian}{\dag}}   & 27.67 & 28.72 & 25.51 & 6.38 & 13.96 & 21.99 & 6.26 & 11.90 &  14.93 & 8 \\

   \noalign{\vskip 1.2mm}
 & \textcolor{denim}{gemini-2.5-pro (srw)}\textsuperscript{\textcolor{greenncs}{\dag}}  & 12.85 & 16.07 & 14.18 & 12.07 & 14.48 & 16.09 &  \textcolor{greenmun}{2.01} &  6.84 &   9.55 & 5 \\
 & \textcolor{denim}{o3 (srw)}\textsuperscript{\textcolor{greenncs}{\dag}}  &  \textcolor{greenmun}{8.19} & 11.06 &  9.22 &  8.47 &  8.04 &  6.42 &  3.38 &  3.54 &   2.36 & 3 \\
 & \textcolor{denim}{sonnet-4 (srw)}\textsuperscript{\textcolor{greenncs}{\dag}}  & 22.72 & 27.95 & 20.13 &  7.66 & 18.64 & 19.42 & 47.88 & 39.52 &  23.35 & 11 \\
& \textcolor{denim}{deepseek-v3.1 (srw)}\textsuperscript{\textcolor{greenncs}{\dag}}   & 14.94 & 21.62 & 16.11 & \textcolor{greenmun}{6.31} & 16.38 & 21.31 & 6.09 & 16.13 &  23.16 & 6 \\
\bottomrule
\end{tabular}
}
\caption{Comparison of relative error percentages for edge estimations. \textbf{Bold} values show best performance among all methods and \textcolor{greenmun}{green} values show best performance among LLMs. Ranking is based on median values. \textsuperscript{\textcolor{carnelian}{\dag}} indicate methods that are difficult to implement in real-world setting, while \textsuperscript{\textcolor{greenncs}{\dag}} methods can be used in real-world tasks.}
\label{tab:edge_est}
\end{table*}

\begin{table*}[ht]
\centering
\resizebox{\linewidth}{!}{%
\begin{tabular}{l ccc ccc ccc ccc ccc c}
\toprule
 & \multicolumn{3}{c}{ego-Twitter} & \multicolumn{3}{c}{twitch-gamers} & \multicolumn{3}{c}{email-EuAll} & \multicolumn{3}{c}{as-skitter} & \multicolumn{3}{c}{wiki-Talk} & Rank \\
\textbf{Method} & Median & Mean & Std & Median & Mean & Std & Median & Mean & Std & Median & Mean & Std & Median & Mean & Std &  \\
\midrule
uniform\textsuperscript{\textcolor{carnelian}{\dag}} & \textbf{1.04} & 1.39 & 0.89 &  \textbf{0.58} &  0.84 & 0.77 &  \textbf{3.90} &  3.45 & 2.26 & \textbf{0.85} & 0.87 & 0.45 & \textbf{1.81} & 2.70 & 2.46 &  1\\
mh\textsuperscript{\textcolor{carnelian}{\dag}} & 18.98 & 18.82 & 1.91 &  21.92 &  22.10 & 0.54 &  30.19 &  30.59 & 1.66 & 28.47 & 28.07 & 1.36 & 22.76 & 22.73 & 0.11 &  2\\
  \noalign{\vskip 1.3mm}
\textcolor{lightcoral}{gemini-2.5-pro (mh)}\textsuperscript{\textcolor{carnelian}{\dag}}  & 16.14 & 28.16 & 22.53 & 35.34 & 51.05 & 32.05 & 36.15 & 35.91 & 16.14 & 37.02 & 43.23 & 19.11 & 49.96 & 144.30 & 237.22 & 3 \\
\textcolor{lightcoral}{o3 (mh)}\textsuperscript{\textcolor{carnelian}{\dag}}              & 55.30 & 128.77 & 190.22 & \textcolor{greenmun}{25.04} & 34.26 & 26.03 & 43.78 & 42.71 & 11.98 & 35.21 & 35.68 & 25.54 & 35.29 & 31.17 & 18.59 & 4 \\
\textcolor{lightcoral}{sonnet-4 (mh)}\textsuperscript{\textcolor{carnelian}{\dag}}        & \textcolor{greenmun}{3.15} & 13.51 & 19.48 & 48.51 & 58.82 & 21.36 & 47.18 & 33.40 & 26.64 & 65.75 & 70.46 & 23.73 & 46.31 & 41.10 & 9.69 & 6 \\
\textcolor{lightcoral}{deepseek-v3.1 (mh)}\textsuperscript{\textcolor{carnelian}{\dag}}        & 59.66 & 45.66 & 30.24 & 74.72 & 55.58 & 35.65 & 32.51 & 34.49 & 12.99 & \textcolor{greenmun}{30.30} & 45.75 & 28.18 & \textcolor{greenmun}{26.88} & 22.31 & 8.08 & 8 \\

\noalign{\vskip 1.2mm}
\textcolor{denim}{gemini-2.5-pro (srw)}\textsuperscript{\textcolor{greenncs}{\dag}}      & 24.41 & 50.94 & 42.41 & 78.64 & 137.64 & 146.08 & 79.58 & 77.03 & 6.46 & 53.59 & 45.73 & 40.59 & 38.48 & 46.46 & 61.45 & 9 \\
\textcolor{denim}{o3 (srw)}\textsuperscript{\textcolor{greenncs}{\dag}}                  & 41.55 & 528.84 & 1106.12 & 36.81 & 27.96 & 15.78 & \textcolor{greenmun}{7.10} & 139.92 & 292.49 & 53.23 & 40.07 & 27.28 & 84.68 & 79.10 & 49.78 & 7 \\
\textcolor{denim}{sonnet-4 (srw)}\textsuperscript{\textcolor{greenncs}{\dag}}            & 25.50 & 30.80 & 33.41 & 47.11 & 52.32 & 43.27 & 26.64 & 33.40 & 28.60 & 53.23 & 56.60 & 31.94 & 44.81 & 37.62 & 24.50 & 5 \\
\textcolor{denim}{deepseek-v3.1 (srw)}\textsuperscript{\textcolor{greenncs}{\dag}}            & 41.64 & 51.07 & 38.57 & 38.87 & 43.44 & 32.19 & 130.34 & 88.83 & 58.71 & 53.60 & 47.82 & 31.02 & 27.66 & 61.59 & 49.67 & 10 \\
\bottomrule
\end{tabular}}
\caption{Comparison of relative error percentages for edge estimations for real-world datasets. \textbf{Bold} values show best performance among all methods and \textcolor{greenmun}{green} values show best performance among LLMs. Ranking is based on median values.}
\label{tab:real_data_edge_est}
\end{table*}

\subsection{Estimation of Influential Nodes}

Table \ref{tab:node_ranking_p50} \& \ref{tab:node_ranking_p100} show the prediction results of node ranking for Precision@50 and Precision@100. The results are similar to Precision@20, with o3 significantly outperforming other LLMs.

\begin{table}[h!]
\centering

\resizebox{\linewidth}{!}{%
\begin{tabular}{lrrr}
\toprule
\textbf{Model} & \textbf{Betweenness} & \textbf{Closeness} & \textbf{PageRank} \\
\midrule
gemini-2.5-pro   & $12.50 \pm 11.80$ & $15.20 \pm 8.91$  & $27.10 \pm 8.80$  \\
o3               & $37.40 \pm 10.02$ & $33.90 \pm 9.31$  & $88.60 \pm 14.90$ \\
sonnet           & $23.10 \pm 11.67$ & $27.10 \pm 12.34$ & $49.90 \pm 24.29$ \\
DeepSeek-V3.1    & $19.80 \pm 14.45$ & $25.10 \pm 15.00$ & $28.10 \pm 10.27$ \\
\bottomrule
\end{tabular}
}
\caption{{Precision@50} (\%) for node centrality values.}
\label{tab:node_ranking_p50}
\end{table}

\begin{table}[h!]
\centering

\resizebox{\linewidth}{!}{%
\begin{tabular}{lrrr}
\toprule
\textbf{Model} & \textbf{Betweenness} & \textbf{Closeness} & \textbf{PageRank} \\
\midrule
gemini-2.5-pro & $17.20 \pm 13.52$ & $16.00 \pm 4.55$  & $28.40 \pm 5.97$  \\
o3             & $43.25 \pm 11.65$ & $35.05 \pm 9.65$  & $85.05 \pm 14.94$ \\
sonnet         & $26.90 \pm 11.09$ & $29.15 \pm 13.68$ & $60.70 \pm 21.17$ \\
DeepSeek-V3.1  & $20.90 \pm 14.60$ & $21.85 \pm 17.40$ & $22.90 \pm 18.80$ \\
\bottomrule
\end{tabular}
}
\caption{{Precision@100} (\%) for node centrality values.}
\label{tab:node_ranking_p100}
\end{table}

\section{\textcolor{black}{Practical Recommendations for Large Graph Analysis with LLMs}}
\color{black}
\label{sec:recommendation}
\subsection{Graph Size Estimation}
\begin{figure}
    \centering
    \includegraphics[width=0.9\linewidth]{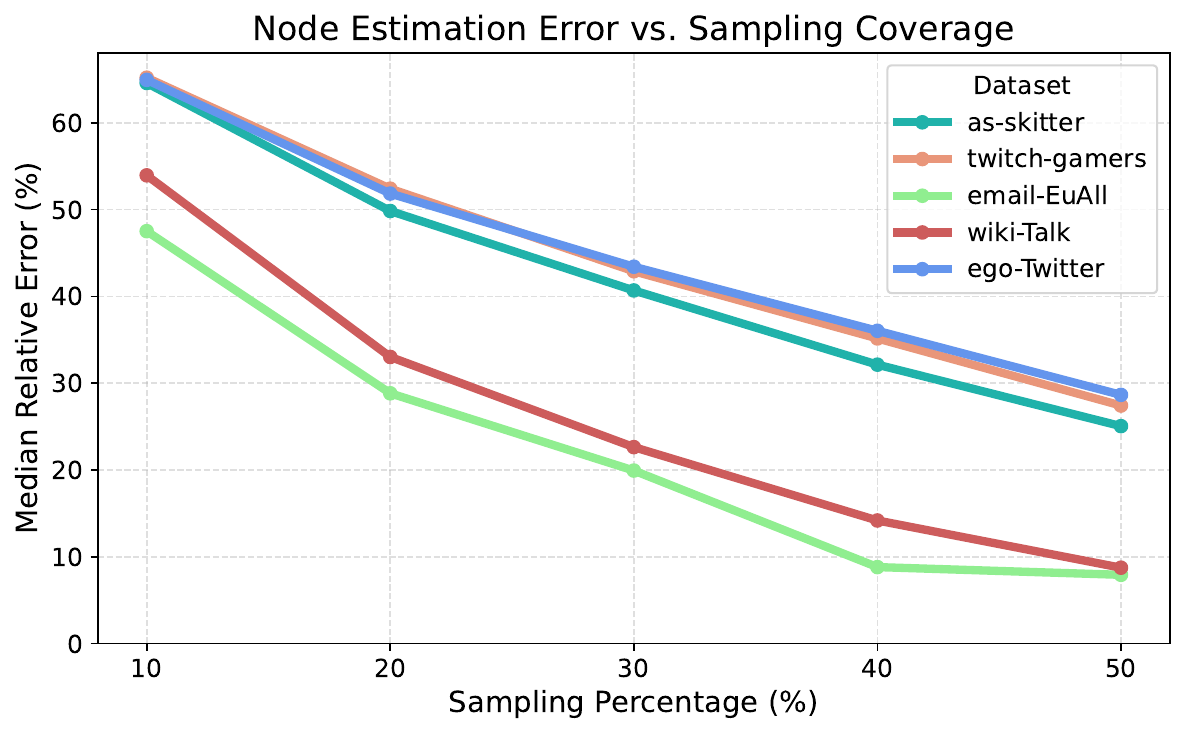}
    \caption{\textcolor{black}{Graph size estimation error with sampling size (srw) for real-world datasets with o3 model.}}
    \label{fig:node_est_change}
\end{figure}

\begin{itemize}
    \item Our results show that LLMs can make reasonable graph size estimates, with a median relative error below 20\% on many graph datasets. 
    \item Unbiased random walks like Metropolis-Hastings (mh-walk) random walk can lead to better performance. However, implementing such walks is costly, especially under limited-access settings. Hence, in practical applications, simple random walks (srw) can be employed with a slight loss in estimation accuracy.
    \item For graph size estimation, the length of random walks should be sufficient to accumulate node collisions. In practice, we found a walk length of at least 10\% of the actual graph size can provide good estimates. However, the optimal length of walk depends on the structure of the graph.
    \item To estimate the number of edges in a graph, LLMs first estimate the number of nodes and the average degree of the graph and calculate the number of edges based on the estimations. Therefore, edge estimations have higher errors compared to node estimations.
    \item Figure \ref{fig:node_est_change} shows that as we increase the length of random walks, node estimation errors by LLMs decrease.
    \item In our experiments, we found that LLMs sometimes hallucinate estimates that are orders of magnitude higher than the ground truth value. Hence, it is advisable to run the analysis a few times, and use the median or some other robust measure to detect and remove outliers.
\end{itemize}

\subsection{Estimate Number of Communities}

\begin{itemize}
    \item Our experiments show that LLMs can detect the presence of multiple communities, with a mean absolute error of $\approx$2 communities (when true count is 5-12), but they are not yet reliable for precise community counts.
    \item For community detection, it is useful to seed random walks from diverse locations in the graph structure to increase approximate coverage. However, in real-world settings, this might be harder to achieve.
    \item Longer walks help to identify communities more easily, as repeated visits to the same nodes within a region provide stronger evidence of clustering. In our experiments, we relied only on the graph structure, however in real-world data, node attributes can be used to bias the walk towards nodes sharing similar attribute types.
    \item While in our experiments, we used a single prompt to detect communities, for larger graphs, walk statistics from different seed nodes can be distributed across multiple prompts, and the resulting community estimates can be aggregated to produce a final prediction.
    
\end{itemize}

\subsection{Estimate Structure of Graph}

\begin{itemize}

    \item LLMs can reliably identify grid-like graphs and distinguish between random and power-law-structured graphs. 
    \item  LLMs use degree distributions to detect graph structure. Using multiple short-to-medium walks from diverse start nodes can help reduce local-region bias.
    \item Estimating graph structure with LLMs can help to select or adapt a probing strategy or downstream algorithm for further graph analysis. It can also support topological characterization real-world graphs for robustness analysis and risk assessment.
\end{itemize}

\subsection{Estimate Influential Nodes in the Graph}
\begin{itemize}
 
    \item For LLMs, identifying PageRank-like importance in graphs is easier, with the o3 model performing the best. PageRank has been used to rank web pages. However, shortest-path based node-ranking measures, such as betweenness centrality and closeness centrality are more difficult to estimate and exhibit higher errors. 
    \item In our experiments, o3 model significantly outperformed other models, therefore, in practical applications, the choice of the LLM model can play a significant role in estimation accuracy.
    \item With stronger performance on PageRank-like ranking measures, LLMs are particularly well suited for social networks or web graphs.
\end{itemize}

\section{Supplementary Method Details}
{\color{black}
\subsection{Graph Sampling with Random Walks}

The following are some methods in the research literature that are often used to uniformly sample large graphs and estimate their properties. We also use these methods as baselines to evaluate the performance of LLMs.

\textit{Uniform Sampling from Nodelist} As our first baseline, we take a simple approach of uniformly sampling nodes from the nodelist of the graph. By using the nodelist, we assume access to the entire graph and use this method to establish the effectiveness of the Chapman Estimator. 

\textit{Metropolis-Hastings sampling} This is an application of the Metropolis-Hastings algorithm for unbiased graph sampling. It uses a modified random walk to draw nodes from the graph such that the transition probabilities of the walk converges into a uniform distribution \cite{metropolis_equation_1953,gjoka_practical_2011}. The transition matrix $P^{MH}$ of the walk is defined as: 


\resizebox{0.95\linewidth}{!}{$
P^{\text{MH}}(u,v) = 
\begin{cases}
    \frac{1}{\max(d(u),d(v))} & \{u,v\} \in E, \\
    1 - \sum_{r\in\Gamma(u)}\frac{1}{\max(d(u),d(r))} & u = v, \\
    0 & \text{otherwise},
\end{cases}
$}

\textit{Max-degree Random Walk Sampling} In this algorithm, to uniformly sample nodes via random walk, the original graph is modified into a regular graph (where all nodes have same degree) by adding self-loops to the nodes so that the degree of each node equals the maximum degree ($d_{max}$) of the original graph. When the walk is at node $u$, the walk selects a neighboring node from $\Gamma(u)$ with a probability of $1/d_{max}$, or stays at node $u$ with the probability of $\frac{d_{max} - d_u}{d_{max}}$. As the walk becomes equivalent to simple random walk on a regular graph, the stationary distribution of the max-degree random walk is a uniform \cite{bar-yossef_approximating_2000,li_random_2015}.

\textit{Return-time Based Estimation} \cite{cooper_fast_2016} proposed a method to estimate global properties of a graph using the times of the first returns of random walks. To estimate the number of nodes, inversely degree-biased weighted random walks are used, where the edge weight is assigned as
\[w(u,v) = \frac{1}{d(u)} + \frac{1}{d(v)}\]
Let $Z(k) = \sum_{i=1}^{k} Z_{i}$ be the time of the $k$th return to the vertex $u$; then the number of nodes can be estimated as 
\[\hat{n} = \frac{Z(k)w(u)}{2k}\]

Under this setting, walks starting from high-weight nodes can exploit structural variations in the graphs to estimate properties efficiently.

\subsection{Community Detection Methods}

Following community detection algorithms are used as baseline in our experiment results:

\textit{Greedy Modularity} This algorithm uses the Clauset-Newman-Moore greedy modularity maximization \cite{clauset_finding_2004} to find the community partition with the largest modularity.

\textit{Louvain Community detection} This is a heuristic method based on modularity optimization. It efficiently uncovers communities in large networks by iteratively aggregating nodes into communities \cite{blondel_fast_2008}.

\textit{Label Propagation} In this algorithm, each vertex in the network is assigned a unique label, which will be used to determine the community it belongs to. Through an iterative process, the community size is expanded as connected groups of nodes adopt the same label \cite{cordasco_community_2010}. 
}

\subsection{Node Ranking}
We use following commonly used node ranking measures:

\textit{Betweenness Centrality} Betweenness centrality measures how often a node lies on the shortest paths between other pairs of nodes. A node with high betweenness can act as a bridge and has strong influence over information flow in the network \cite{brandes_centrality_2007, borassi_kadabra_2019, maurya_graph_2021}. For a given node $v$, it is defined as:

    \[ C_B(v) = \sum_{s \neq v \neq t} \frac{\sigma_{st}(v)}{\sigma_{st}} \]

where $\sigma_{st}$ is the total number of shortest paths from node $s$ to node $t$, and $\sigma_{st}(v)$ is the number of those paths that pass through $v$.

\textit{Closeness Centrality} Closeness centrality measures how near a node is to all other nodes in the graph based on shortest-path distances. Nodes with high closeness can quickly interact with or spread information to the rest of the graph or network \cite{beauchamp_improved_1965,freeman_centrality_1978}. For a given node $v$ in a graph, it is defined as:

\[ C_C(v) = \frac{n-1}{\sum_{u \neq v} \delta(v, u)} \]
where $\delta(v, u)$ is the shortest path distance between nodes $v$ and $u$.

\textit{PageRank} PageRank assigns an importance score to each node based on the principle that a node is important if it is linked to by other important nodes. Originally designed for web page ranking \cite{brin_anatomy_1998, gleich_pagerank_2015}, it has been widely adopted in other fields as well. 

\[PR(v) = \frac{1-\lambda}{n} + \lambda \sum_{u \in \Gamma(v)} \frac{PR(u)}{d(u)}\]
where $\lambda \in (0,1)$ is the damping factor, $\Gamma(v)$ denotes 
the set of neighboring nodes of $v$, and $d(u)$ is the degree of node $u$.

\section{\textcolor{black}{Graph Datasets for LLM Reasoning}}
\textcolor{black}{Many graph benchmarks have been proposed to evaluate the reasoning capabilities of LLMs. Table \ref{tab:graph_datasets} shows the statistics of the sizes of the graphs (number of nodes) present in these datasets. As each dataset has many graphs, we present the maximum or the average number of nodes as reported in the respective papers.}

\textcolor{black}{Table \ref{tab:benchmark_dataset} shows of the statistics of the datasets in the benchmark \texttt{EstGraph}. While we provide a large number of graphs in the datasets for benchmarking, similar to \texttt{NLGraph} paper, we use a smaller subset for evaluations in our experiments due to API costs. Table \ref{tab:graph_datasets} shows that the sizes of the graphs used in our paper are many orders of magnitude larger than those in other graph datasets. However, our paper focuses on estimation tasks that are relevant to very large graphs, where access to the whole graph is infeasible; these tasks are not applicable to very small graphs. Therefore, the range of tasks evaluated by other graph benchmarks is different from ours and are not directly comparable. Combining our approach with other graph benchmarks can provide comprehensive evaluation strategy for LLM reasoning on graph-structured data.}

\begin{table}[h!]
\centering
\resizebox{\linewidth}{!}{%
\begin{tabular}{l c}
\toprule
\textbf{Dataset} & \textbf{Graph Size} \\
\midrule
NLGraph \cite{wang_can_2023}    & Maximum: 20          \\
GraphQA \cite{fatemi_talk_2024}   & Average: 12.3        \\
GraphPattern \cite{dai_how_2025}   & Average: 29.5 \\
GraphArena \cite{tang_grapharena_2025} & Maximum: 50          \\
\midrule
\textbf{Ours} (synthetic) & Maximum: 100,000 \\
\textbf{Ours} (real-world) & Maximum: 2,388,953 \\
\bottomrule
\end{tabular}
}
\caption{Size of graph datasets for LLM Reasoning. Maximum or average number of nodes in the graph dataset is reported.}
\label{tab:graph_datasets}
\end{table}


\begin{table}[t]
\centering

\resizebox{\linewidth}{!}{%
\begin{tabular}{@{} l l c @{}} 
\toprule
\textbf{Estimation Task} & \textbf{Graph Type} & \textbf{Number of Graphs} \\
\midrule
\multirow{9}{*}{Graph Size Estimation}
  & BA (small)   & 100 \\
  & BA (medium)  & 100 \\
  & BA (large)   & 100 \\
  & ER (small)   & 100 \\
  & ER (medium)  & 100 \\
  & ER (large)   & 100 \\
  & GRP (small)  & 100 \\
  & GRP (medium) & 100 \\
  & GRP (large)  & 100 \\
\midrule
Number of communities & LFR  & 100 \\
\midrule
\multirow{4}{*}{Structure Prediction}
  & ER           & 100 \\
  & BA           & 100 \\
  & Grid         & 100 \\
  & LFR          & 100 \\
\midrule
Node Ranking & LFR & 100 \\
\bottomrule
\end{tabular}
}
\caption{Dataset statistics for \texttt{EstGraph} benchmark}
\label{tab:benchmark_dataset}
\end{table}

\section{\textcolor{black}{Experiment Settings}}
\color{black}
\label{sec:appendix_experiment_parameters}
In this section, we provide the details of the experiment settings related to each task described in Section \ref{sec:est_tasks}:

\subsection{Evaluated Models}
For our experiments, we used the following three reasoning LLMs:
\begin{itemize}
    \item gemini-2.5-pro from Google

    \item sonnet-4 from Anthropic 

    \item o3 from OpenAI (reasoning effort: high)

    \item DeepSeek-V3.1 (Thinking Mode)
\end{itemize}

We generated prompts for each task and used the API endpoints of the respective models. \textcolor{black}Source code and graph data used in this experiment is available at \url{https://zenodo.org/records/19632942}.

\subsection{Estimating Number of Nodes and Edges}
For this task used synthetic as well as real-world graph datasets. Details of the datasets are as follows:

\subsubsection{Dataset Generation}
\textit{Synthetic Graph Generation} We generated three types of synthetic graphs covering a wide range of  structural characteristics, using the \texttt{NetworkX} framework \cite{hagberg_exploring_2008}\footnote{https://networkx.org/}. Table \ref{tab:synthetic_graph} provides the \texttt{NetworkX} functions used to generate the random graphs. We also used \texttt{NetworKit\footnote{https://networkit.github.io/}} \cite{angriman_algorithms_2022} for graph processing in some experiments. For each graph type and size, we generated synthetic graphs to create the dataset. The graphs are divided into three size categories: Small (100–1,000 nodes), Medium (1,000–10,000 nodes), and Large (10,000–100,000 nodes).

\begin{table*}[h!]
\centering
\resizebox{\linewidth}{!}{%
\begin{tabular}{|l|l|l|}
\hline
\textbf{Graph Type} & \textbf{NetworkX Generating Function} & \textbf{Generation parameters} \\
\hline
Barabasi-Albert Graph (BA) & \texttt{barabasi\_albert\_graph()} & $m=random.randint(3,5)$ \\
Erdos-Renyi Random Graph (ER) & \texttt{fast\_gnp\_random\_graph()} & number of edges 5-10 times $|N|$ \\
Gaussian Random Partition Graph (GRP) & \texttt{gaussian\_random\_partition\_graph()} & $s=random.uniform(0.05,0.2)\cdot |N|$; $v=\frac{s}{2}$  \\
\hline
\end{tabular}
}
\caption{NetworkX functions used for generating synthetic graphs.}

\label{tab:synthetic_graph}
\end{table*}

\textit{Real-World Datasets} We use 5 real-world datasets from the SNAP graph repository \cite{leskovec_snap_2016}\footnote{https://snap.stanford.edu/snap/}. 

\begin{table*}[h!]
\centering
\resizebox{0.7\linewidth}{!}{%
\begin{tabular}{|l|c|c|c|}
\hline
\textbf{Dataset Name} & \textbf{Number of Nodes} & \textbf{Number of Edges} & \textbf{Dataset Type} \\
\hline
as-skitter & 1,694,616 & 11,094,209 & Autonomous system graph \\
twitch-gamers & 168,114 & 6,797,557 & Social network \\
email-EuAll & 224,832 & 340,795 & E-mail network \\
wiki-Talk & 2,388,953 & 4,656,682 & Communication network \\
ego-Twitter & 81,306 & 1,342,310 & Social network \\
\hline
\end{tabular}
}
\caption{Summary of Real-world datasets}
\end{table*}

\subsubsection{Prompt generation}
\begin{figure*}[h]
    \centering
    \includegraphics[width=0.9\textwidth]{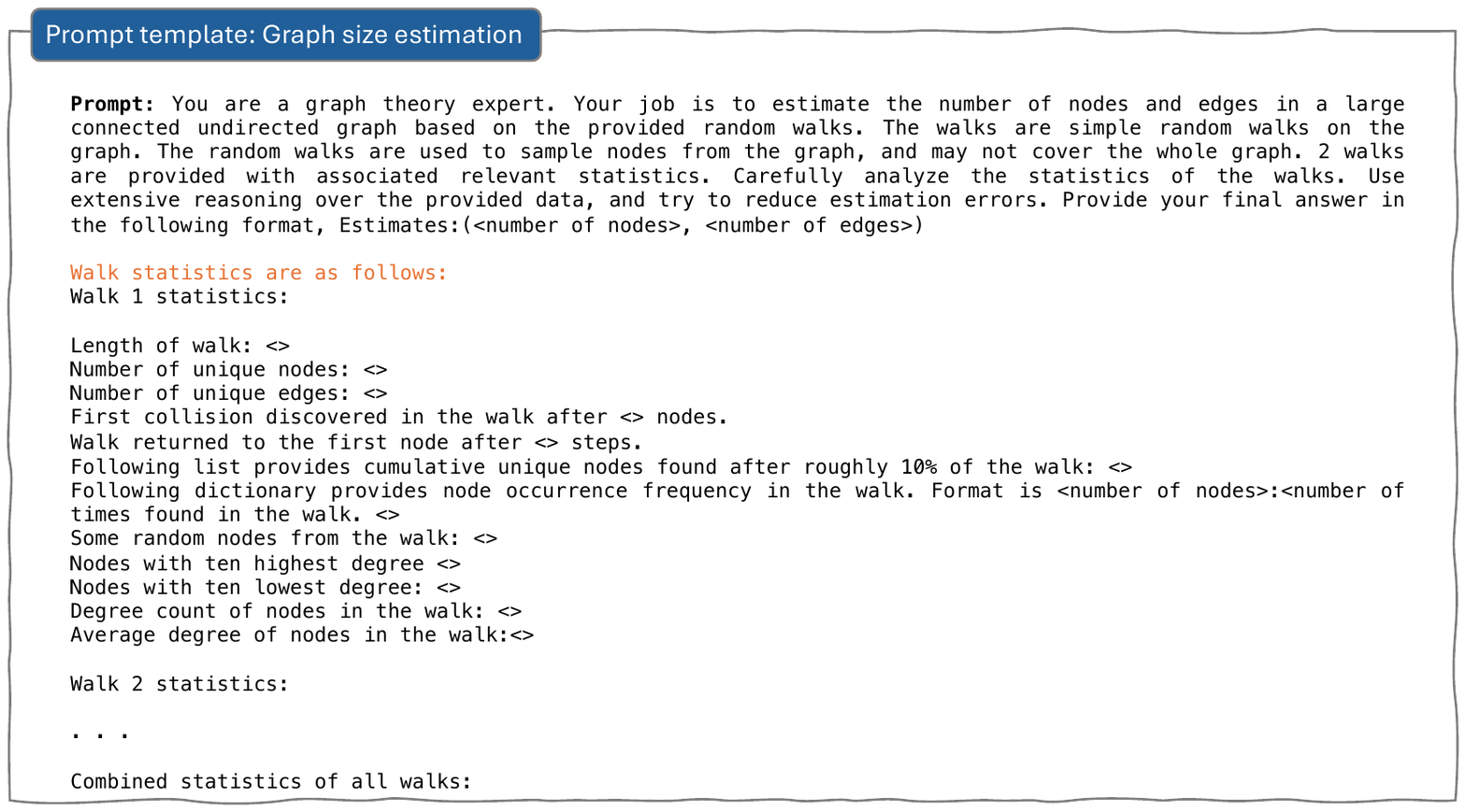}
    \vspace{-8mm}
    \caption{\textcolor{black}{Figure shows the prompt template for estimating number of nodes and edges in the graphs.}}
    \vspace{-5mm}
    \label{fig:prompt_template}
\end{figure*}

We chose the minimum sample size to be 20\% of the graph size because, for the uniform sampling baseline, a lower sample size would not yield a sufficient number of overlapping nodes for the estimation calculations. For the other methods as well, we set the sampling size to 20\% of the graph size. For the random walks on the graph, we set a \textit{burn-in} period of 10\% of the graph size. A \textit{burn-in period} is often included in graph estimation tasks during which the nodes obtained during the walk are discarded. This avoids any bias in estimations due to the choice of the initial source node.
In the case of the \texttt{return\_walk} method, we set the number of returns to 10. The prompt template for this task is shown in Figure \ref{fig:prompt_template}.

In the prompt, we provide the following statistics of the walk:
\begin{itemize}

    \item \textcolor{black}{Length of the walk (provides the basic information of sampled nodes)}

    \item \textcolor{black}{Number of unique nodes (provides the basic information of sampled nodes)}

    \item \textcolor{black}{Number of unique edges (provides the basic information of sampled nodes)}

    \item \textcolor{black}{Time to first node collision during the walk (provides a hint of neighborhood structure or local density of the graph and how likely a walk encounters repeated nodes.)}

    \item \textcolor{black}{Number of steps when the walk returned to the initial source node (provides a hint of neighborhood structure or local density of the graph and how likely a walk encounters the source node.)}

    \item \textcolor{black}{Number of new nodes discovered for every 10\% increment of the walk (provides a hint of extent of local clustering in the graph near the source node of the walk)}

    \item \textcolor{black}{Names of some of the nodes. To avoid direct leakage of node count information, we anonymized the nodes by adding a large random number to each node name. This setting mimics real-world datasets where node names are available as serial numbers.}

    \item \textcolor{black}{Nodes with the ten highest degrees in the walk (provides limited information whether high degree nodes are clustering together)}

    \item \textcolor{black}{Nodes with the ten lowest degrees in the walk (provides limited information whether low degree nodes are clustering together)}

    \item \textcolor{black}{Degree distribution of the walk (provides information on degree distribution of the walk)}

    \item \textcolor{black}{Average degree of the nodes in the walk (provides the basic information of sampled nodes)}
\end{itemize}

We also provide a statistics of all walks combined at the end. Many of these statistics are used in the graph literature to study the characteristics of the graphs. 

\subsection{Estimating Number of Communities}
\subsubsection{Dataset Generation}
For this task, we randomly generated LFR benchmark graphs with size ranging from 1000 to 5000 nodes. The number of communities ranged from 5 to 12. In community detection tasks, the walks originating in a community should be long enough to traverse through the community and detect repetitions of nodes in the walk. To detect all communities, we also need walks originating in every communities. Therefore, we set the maximum size of the graph dataset to 5,000 nodes.

\subsubsection{Prompt Generation}
\begin{figure*}[h]
    \centering
    \includegraphics[width=\linewidth]{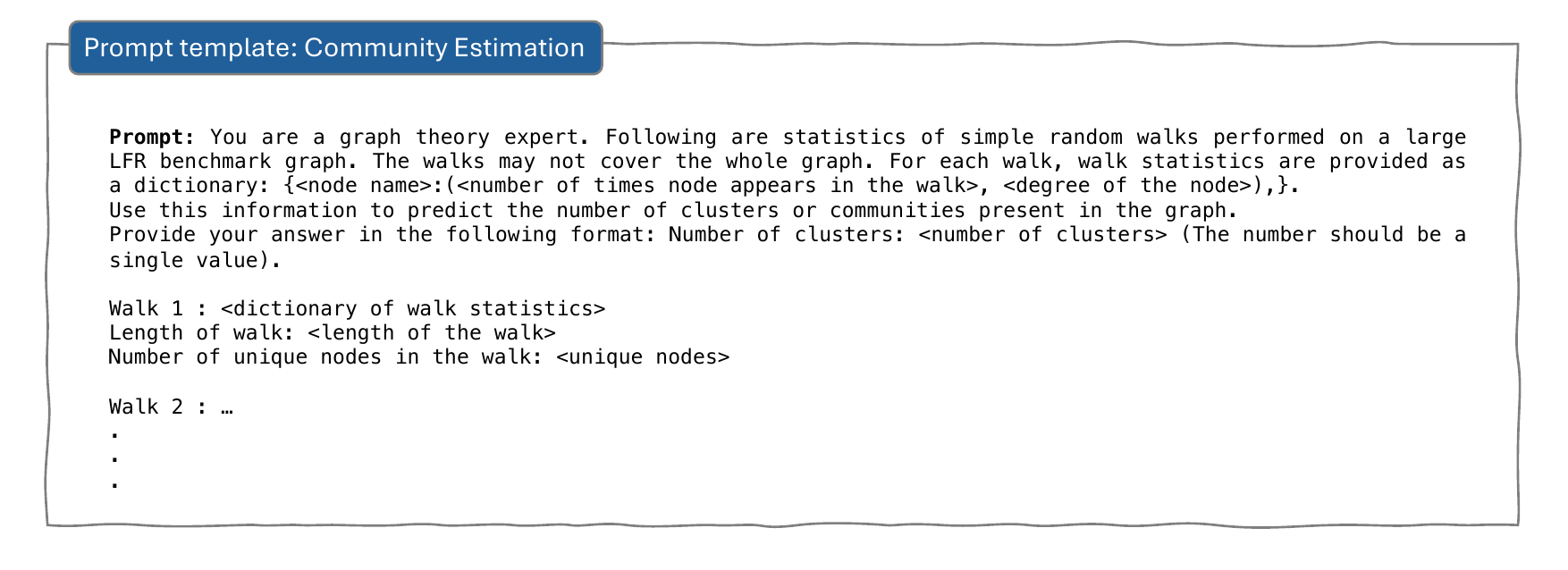}
    \caption{\textcolor{black}{Prompt template for community estimation}}
    \label{fig:community_est}
\end{figure*}

To provide sufficient information to the LLMs, we select two or three initial nodes from each community based on the ground truth. The length of each simple random walk is set to 300, which can cover most community sizes. For each walk, we record the number of unique nodes, how many times each node was visited, and the degree of the nodes. We create a dictionary with nodes as keys and their number of visits and degrees as values. This process is repeated for all walks and the resulting data is added to the input prompt for the LLMs. Figure \ref{fig:community_est} shows the prompt template to create input prompts for the LLMs.

\subsection{Estimate Graph Structure}
\subsubsection{Dataset Creation}
For the task of predicting graph structures from random walks, we created synthetic datasets of four types of graphs: BA, ER, LFR, and Grid graphs. For the BA and ER graph datasets, we reused the medium-sized graphs generated in the previous task. For the LFR and Grid graphs, we generated graphs of up to 10,000 nodes. For grid graphs, we generated three types: hexagonal lattice graphs, triangular lattice graphs, and hypercube graphs using \texttt{NetworkX} generator functions.

\subsubsection{\textcolor{black}{Degree Distribution of Graphs}}
\begin{figure*}
    \centering
    \includegraphics[width=0.7\linewidth]{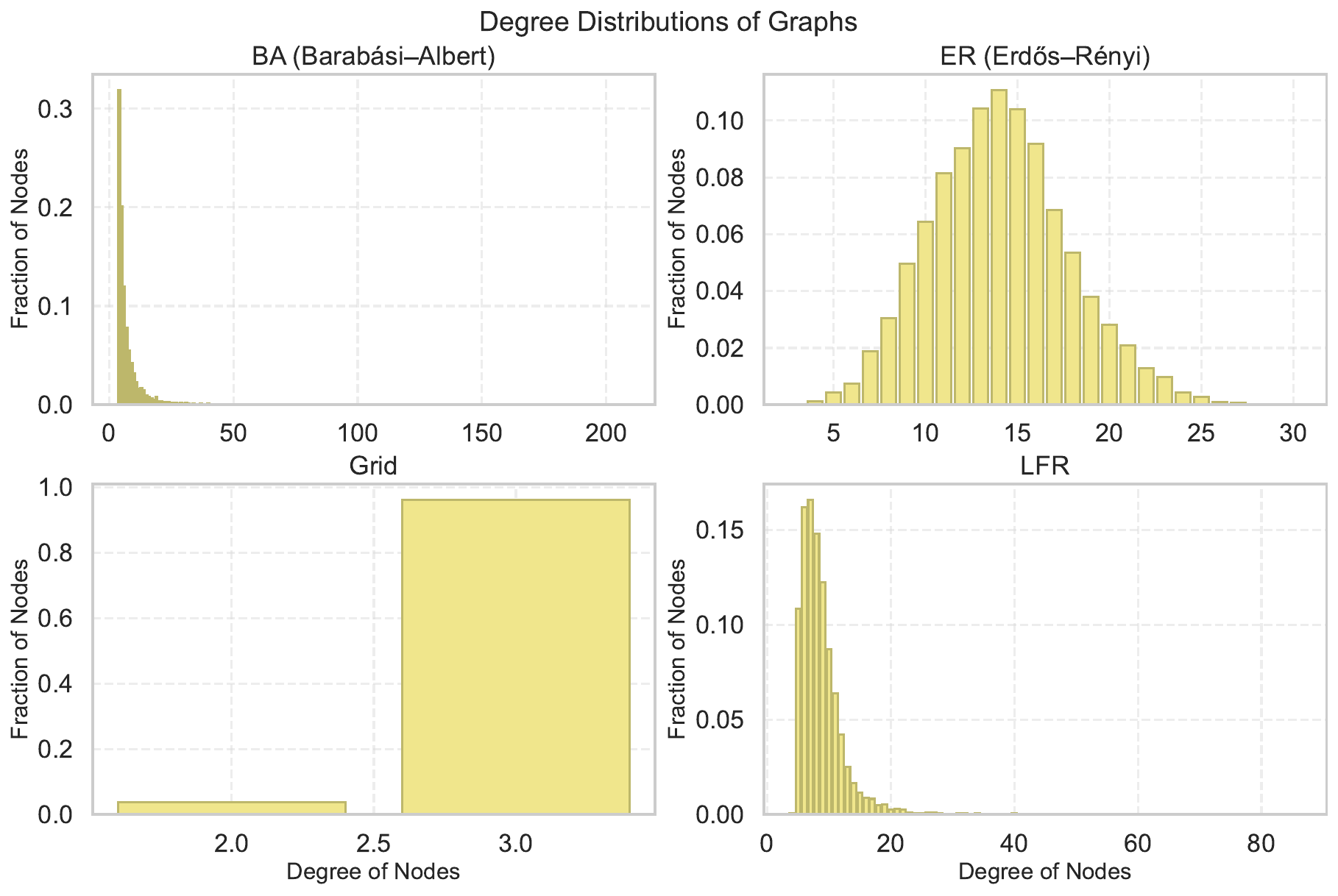}
    \caption{\textcolor{black}{Figure shows the degree distribution of four types of graphs: BA, ER, Grid and LFR}}
    \label{fig:degree_distribution}
\end{figure*}

\textcolor{black}{Figure \ref{fig:degree_distribution} shows the degree distribution of graph types used in this task. The figure illustrates that the node degrees of the ER graph are spread over a wide range of values. In contrast, for BA/LFR graphs, the degree distribution is skewed, with a few nodes having very large degrees and most other nodes having small degrees. Although both LFR and BA graphs have skewed degree distributions, BA graphs (also known as power-law or scale-free graphs) exhibit significantly higher skewness than LFR graphs. When we provide degree information for various nodes in prompts to LLMs, they can distinguish between ER graphs and BA/LFR graphs.}

\subsubsection{Prompt Generation}
\begin{figure*}
    \centering
    \includegraphics[width=\linewidth]{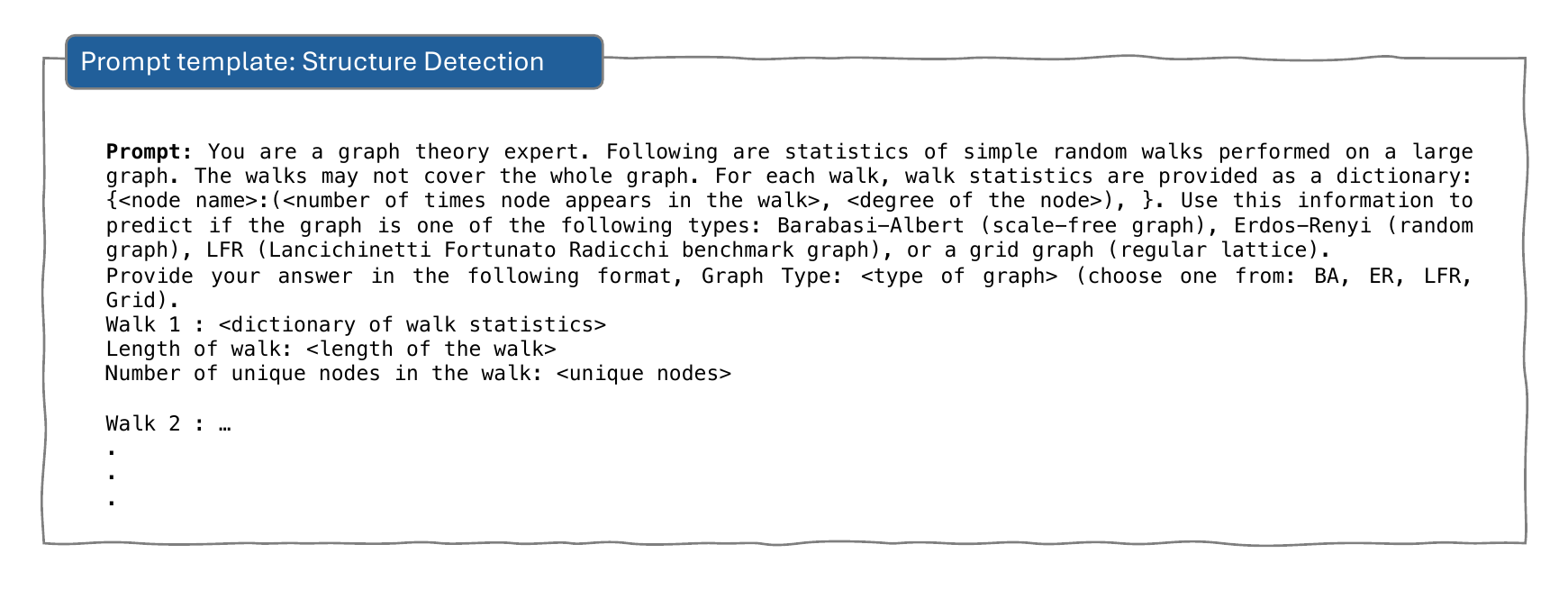}
    \caption{\textcolor{black}{Prompt template for the graph structure prediction }}
    \label{fig:prompt_structure_est}
\end{figure*}

To estimate the structure of a graph, random walks provide crucial information about the degree distribution. For example, in power-law graphs such as BA, the degree distribution is highly skewed, with a few nodes having very high degrees, while ER and grid graphs typically exhibit a uniform node degree distribution. For prompt construction, we perform five walks of length equal to 10\% of the number of nodes in the graph. The initial seed nodes are chosen randomly. For each walk, we record the number of unique nodes, how many times each node was visited, and the degree of the nodes. We create a dictionary with nodes as keys and their number of visits and degrees as values. Figure \ref{fig:prompt_structure_est} shows the template used to create input prompts for this task.

\subsection{Estimating Node Ranking}
\subsubsection{Dataset Creation}
We evaluate top-$k$ node ranking on LFR graphs with community structure. For this task, we randomly generated LFR benchmark graphs with size ranging from 1000 to 5000 nodes. The number of communities ranges from 5 to 12. Node ranking measures like Betweenness Centrality, Closeness Centrality and PageRank were computed for all graphs and used to measure performance of LLMs.

\subsubsection{Prompt Generation}
For prompt construction in the node ranking estimation task, we provide graph statistics similar to those used in the previous task of estimation of graph structure.

\vspace{-3cm}

\begin{figure*}
    \centering
    \includegraphics[width=\linewidth]{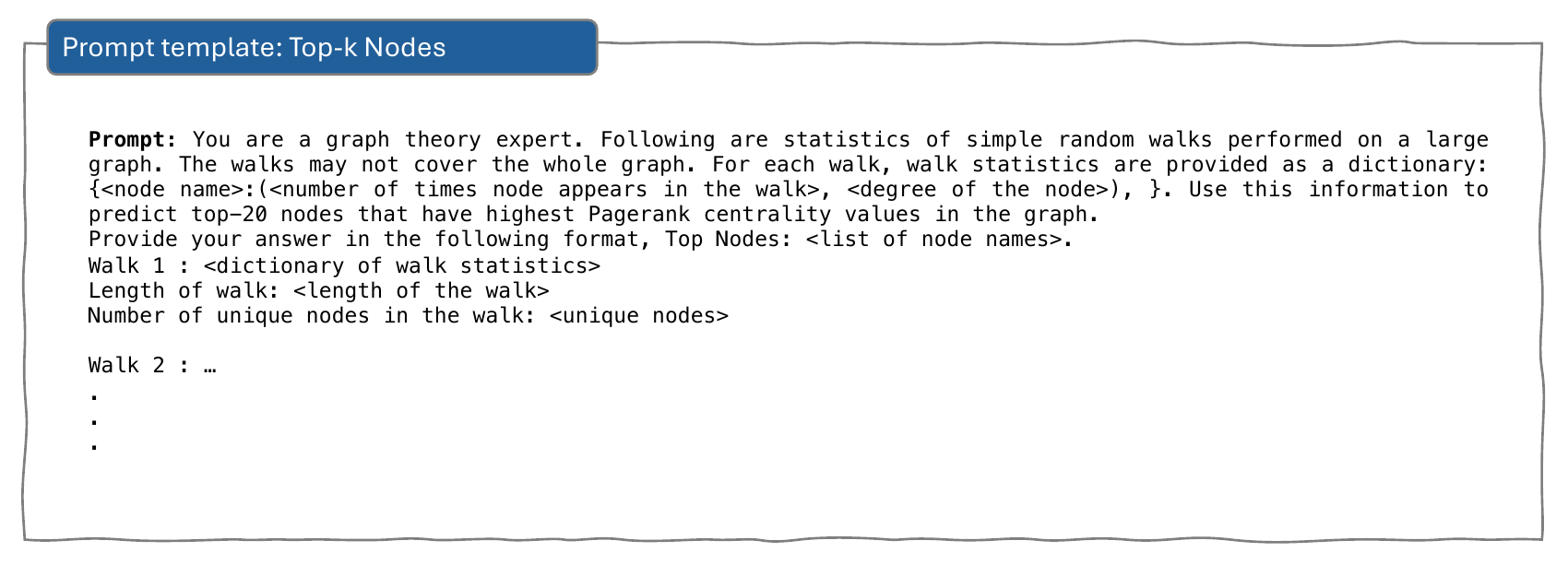}
    \caption{\textcolor{black}{Prompt template for the top-k node prediction }}
    \label{fig:prompt_template_topk}
\end{figure*}




\setlength{\fboxsep}{0pt}     
\setlength{\fboxrule}{1pt}

\begin{figure*}
    \centering
    \includegraphics[width=\linewidth,frame]{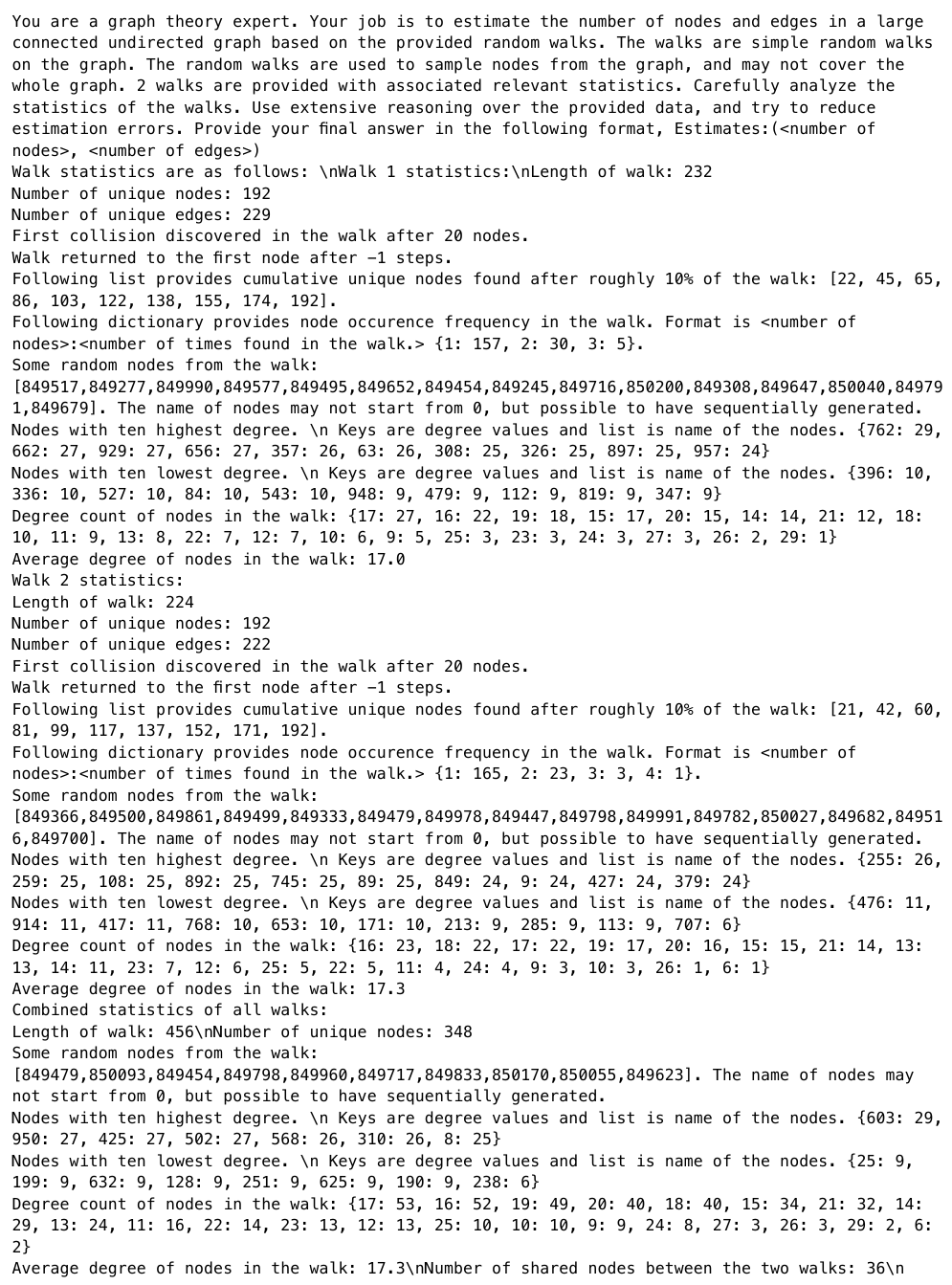}
    \caption{Prompt example for graph size estimation}
    \label{fig:prompt_graph_size_est}
\end{figure*}

\begin{figure*}
    \centering
    \includegraphics[width=\linewidth,frame]{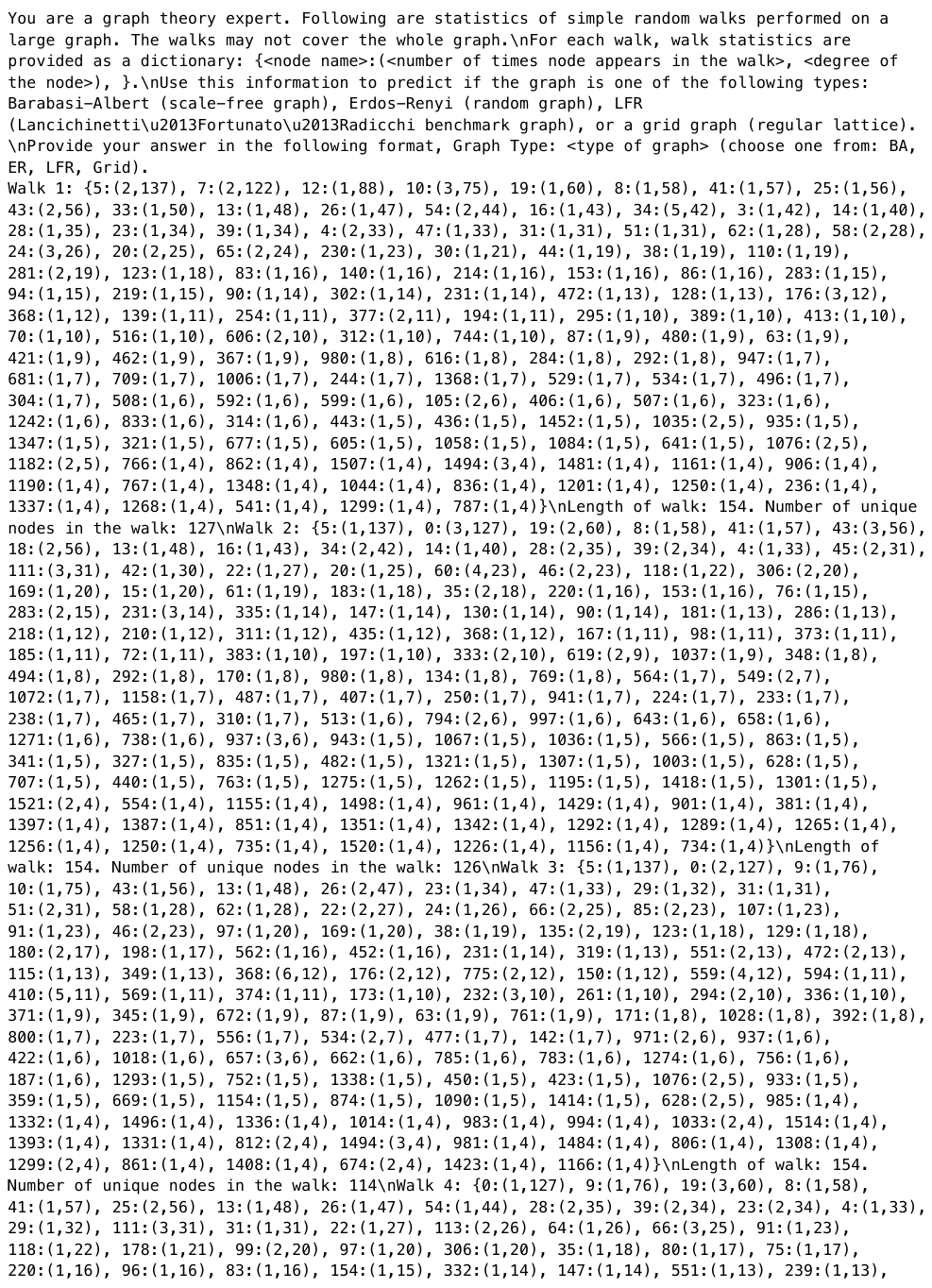}
    \caption*{Prompt example for graph structure prediction (continued)}
\end{figure*}

\begin{figure*}
    \centering
    \includegraphics[width=\linewidth,frame]{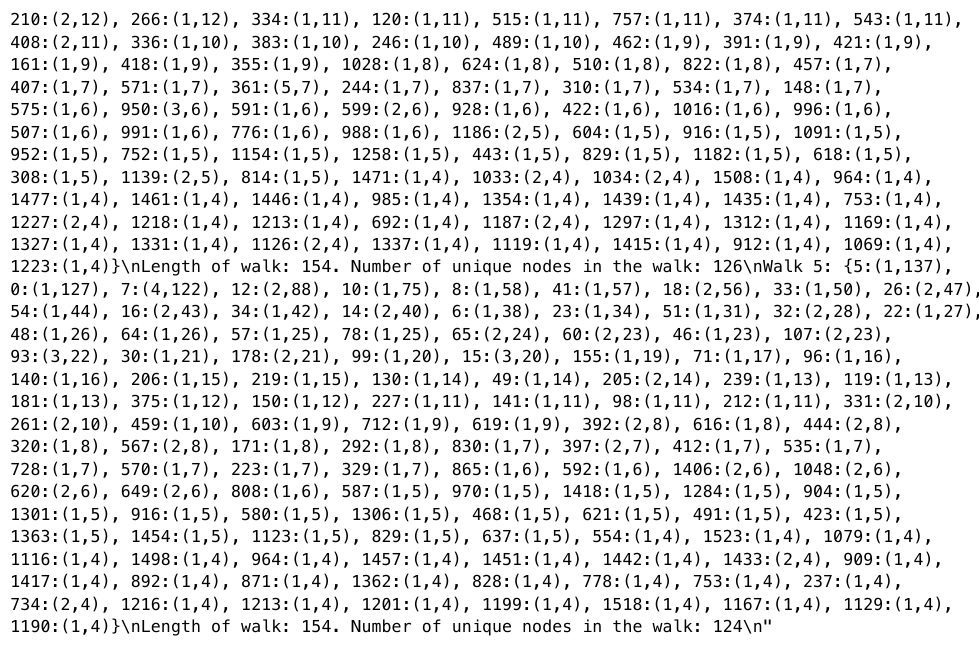}
    \caption{Prompt example for graph structure prediction}
    \label{fig:prompt_graph_struct_pred}
\end{figure*}

\begin{figure*}
    \centering
    \includegraphics[width=\linewidth,frame]{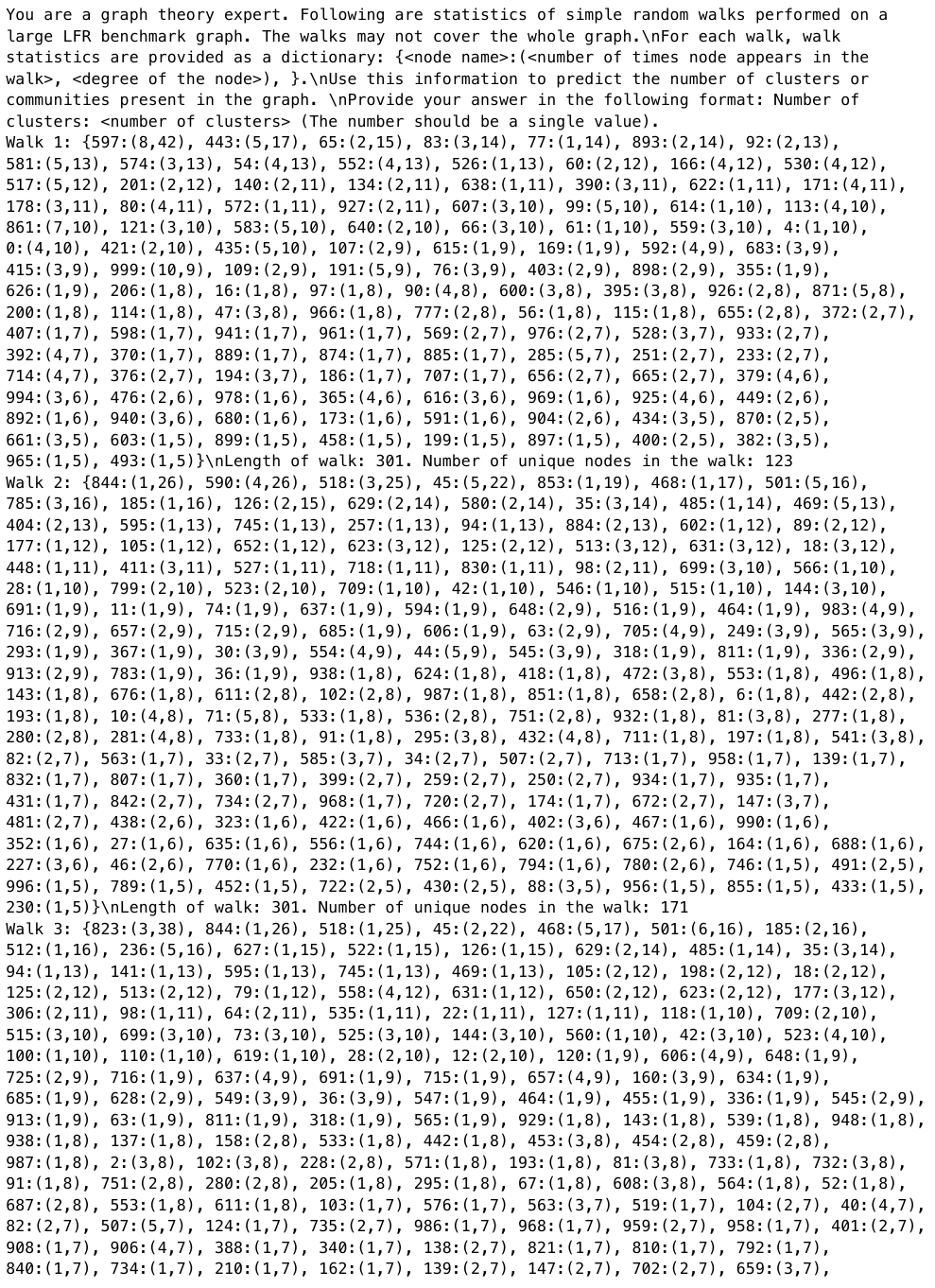}
    \caption*{Prompt example for number of clusters prediction (continued)}
\end{figure*}

\begin{figure*}
    \centering
    \includegraphics[width=\linewidth,frame]{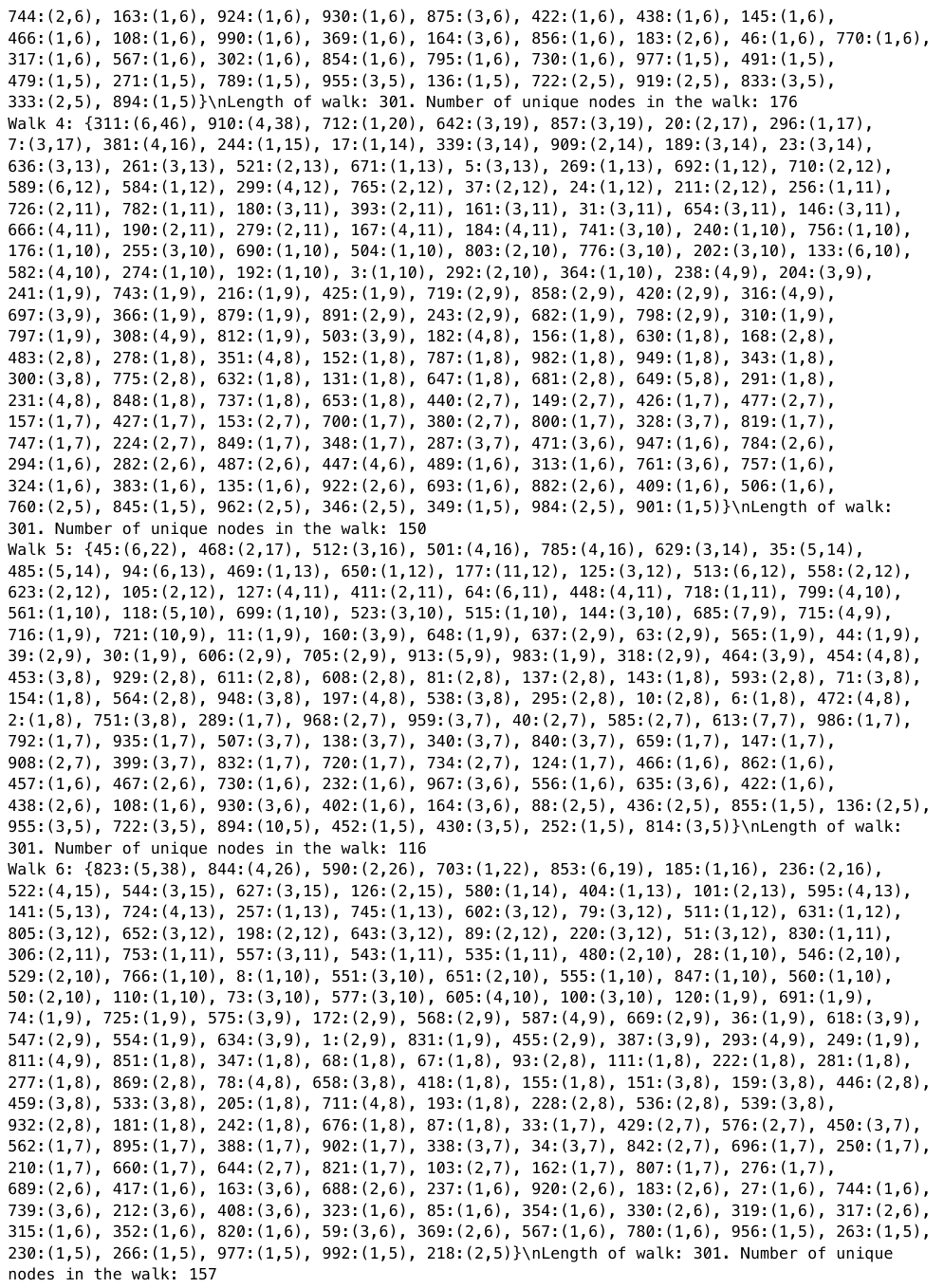}
    \caption*{Prompt example for number of clusters prediction (continued)}
\end{figure*}

\begin{figure*}
    \centering
    \includegraphics[width=\linewidth,frame]{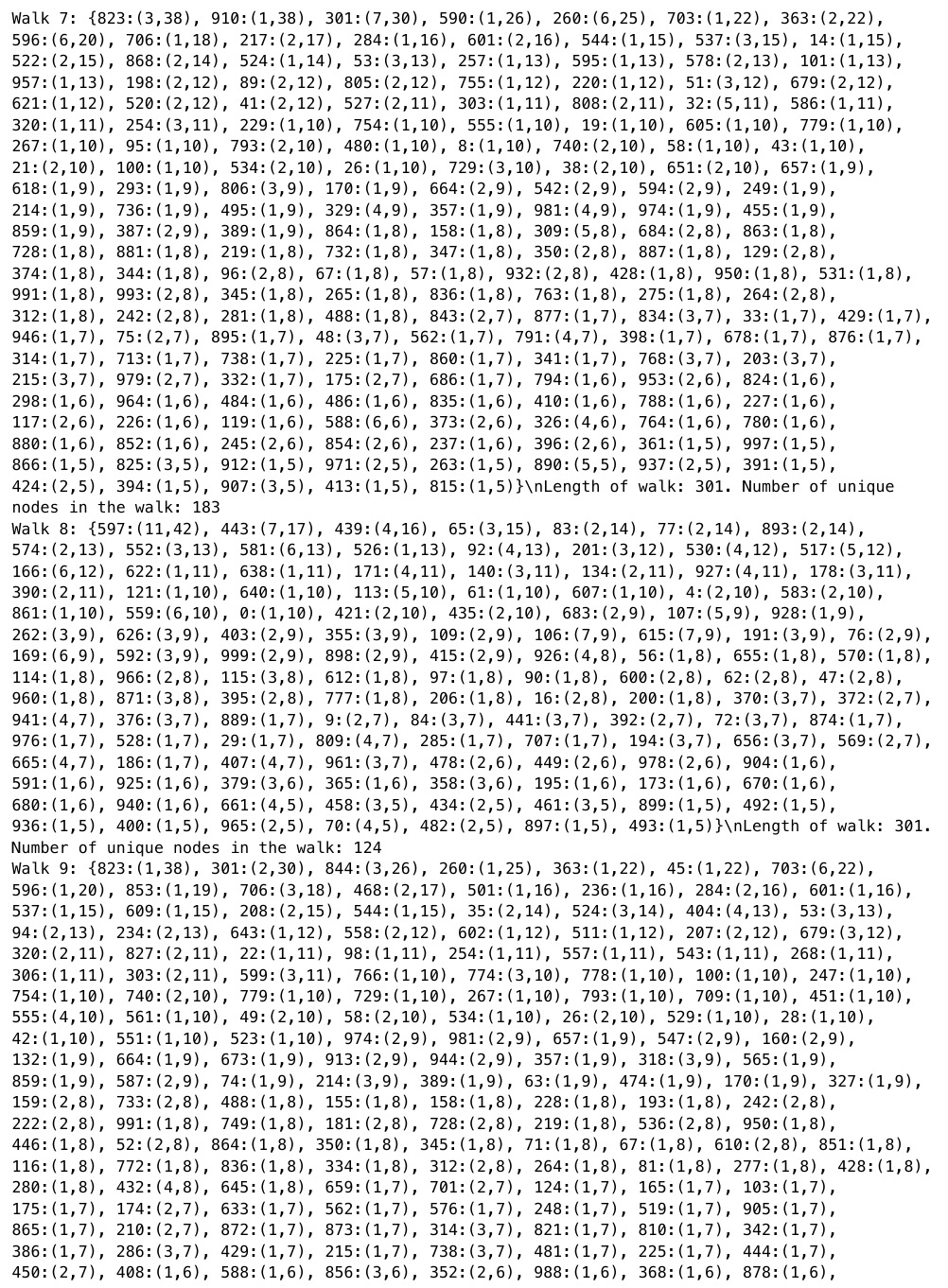}
    \caption*{Prompt example for number of clusters prediction (continued)}
\end{figure*}

\begin{figure*}
    \centering
    \includegraphics[width=\linewidth,frame]{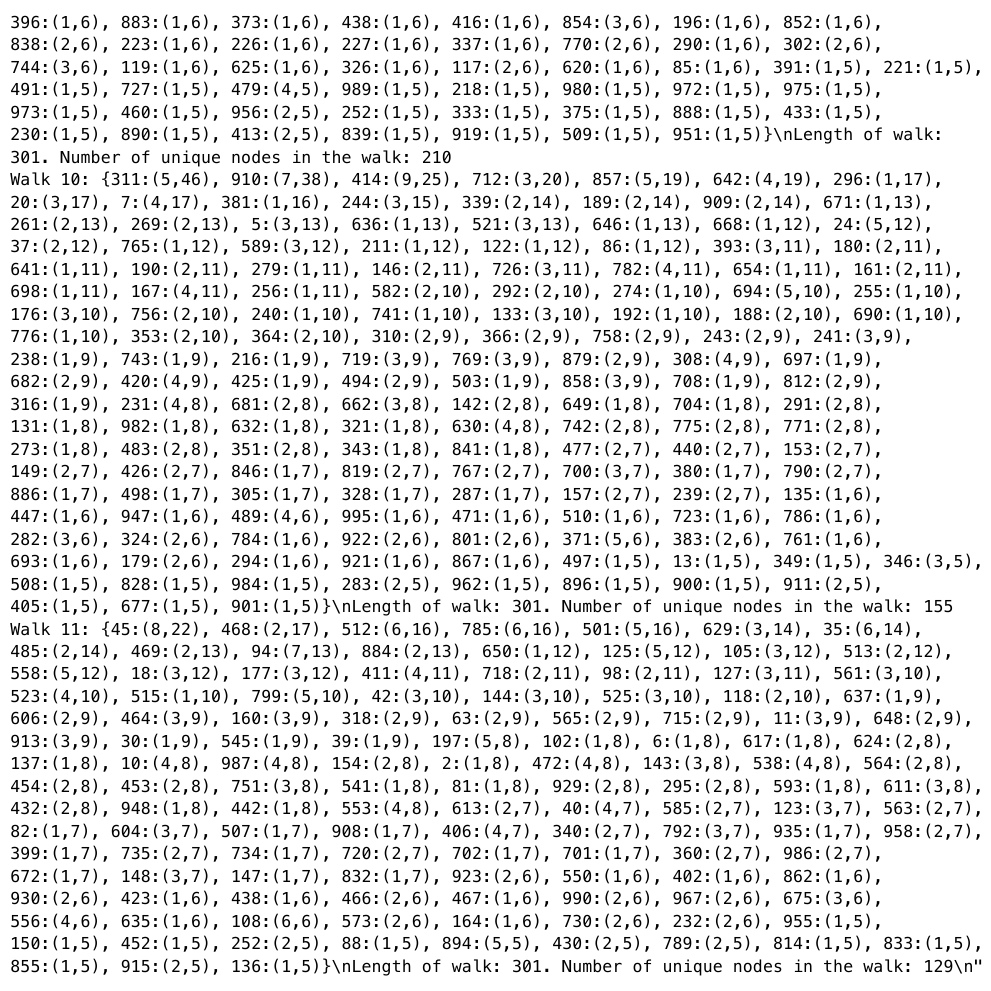}
    \caption{Prompt example for number of clusters prediction}
    \vspace{5cm}
    \label{fig:prompt_cluster_pred}
\end{figure*}

\begin{figure*}
    \centering
    \includegraphics[width=\linewidth,frame]{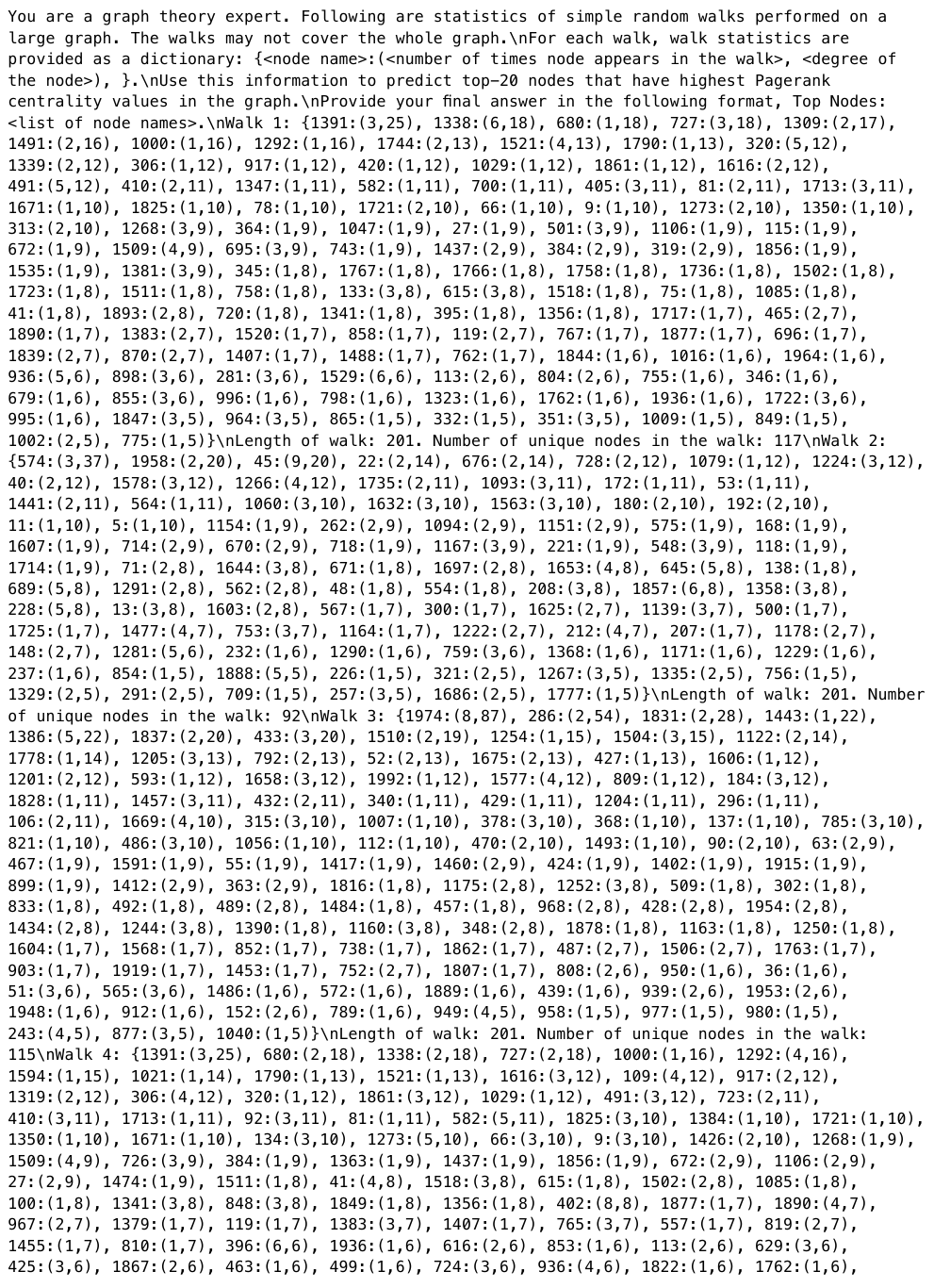}
    \caption*{Prompt example for top-20 Pagerank centrality nodes prediction (continued)}
\end{figure*}

\begin{figure*}
    \centering
    \includegraphics[width=\linewidth,frame]{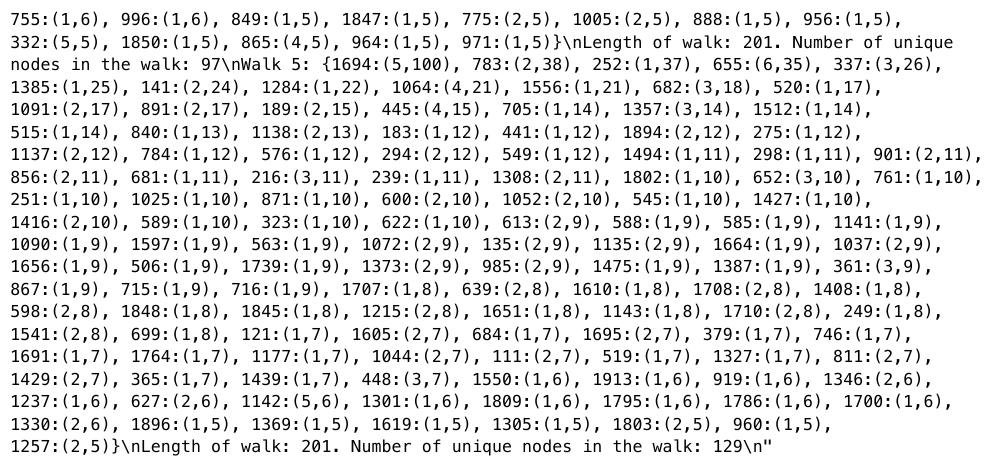}
    \caption{Prompt example for top-20 Pagerank centrality nodes prediction}
    \vspace{5cm}
    \label{fig:prompt_topk}
\end{figure*}

\begin{figure*}
    \centering
    \includegraphics[width=\linewidth,frame]{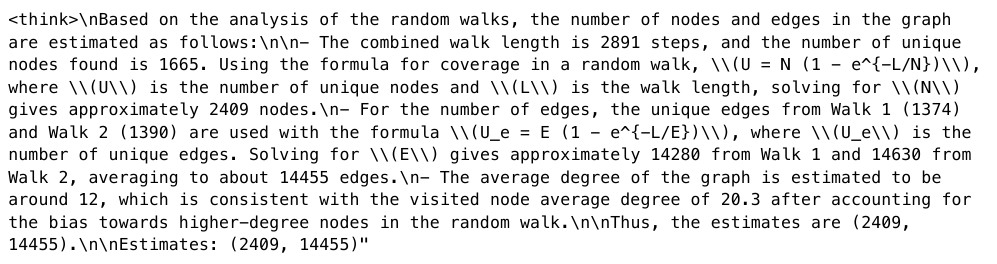}
    \caption{Response example from DeepSeek-V3.1 for the task of graph size estimation. Due to the large amount of text, the reasoning portion of the response is not included.}
    \label{fig:response_graph_est}
\end{figure*}

\begin{figure*}
    \centering
    \includegraphics[width=\linewidth, frame]{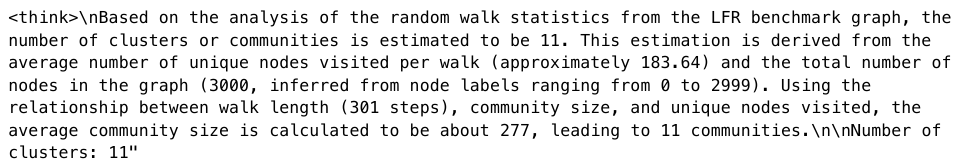}
    \caption{Response example from DeepSeek-V3.1 for the task of prediction of number of communities. The actual number of communities in the input graph are 10. Due to the large amount of text, the reasoning portion of the response is not included.}
    \label{fig:response_num_cluster}
\end{figure*}

\begin{figure*}
    \centering
    \includegraphics[width=\linewidth, frame]{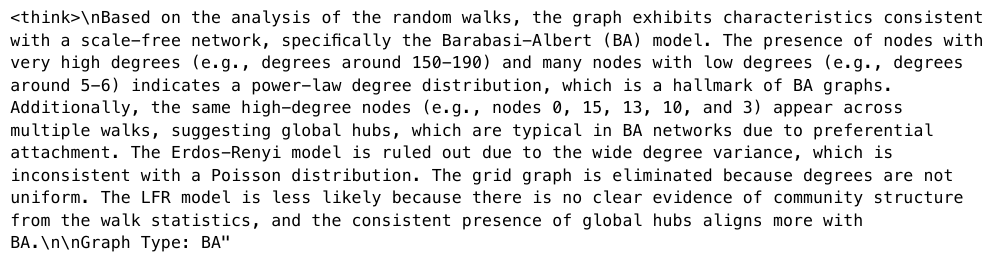}
    \caption{Response example from DeepSeek-V3.1 for the task of graph structure prediction. The actual graph-type for the input graph is BA. Due to the large amount of text, the reasoning portion of the response is not included.}
    \label{fig:response_est_structure}
\end{figure*}

\end{document}